\pdfoutput=1
\documentclass[journal]{IEEEtran}

\usepackage{graphicx}
\usepackage{amsmath}
\usepackage{amssymb}
\usepackage{subcaption}
\usepackage{cite}
\usepackage[hidelinks]{hyperref}


\hypersetup{
    pdftitle={Self-Motivated Growing Neural Network for Adaptive Architecture via Local Structural Plasticity},
    pdfauthor={Yiyang Jia and Chengxu Zhou}
}

\begin{document}

\title{Self-Motivated Growing Neural Network for Adaptive Architecture via Local Structural Plasticity}

\author{Yiyang~Jia,~Chengxu~Zhou%
\thanks{The authors are with the Department of Computer Science, University College London, United Kingdom (\tt\small yiyang.jia.24@ucl.ac.uk; chengxu.zhou@ucl.ac.uk).}%
}

\markboth{arXiv preprint}%
{Jia and Zhou: Self-Motivated Growing Neural Network for Adaptive Architecture via Local Structural Plasticity}

\maketitle

\begin{abstract}
Control policies are often implemented with fixed-capacity multilayer perceptrons trained by backpropagation, which require architecture selection in advance and cannot adapt their capacity during learning. This paper introduces the Self-Motivated Growing Neural Network (SMGrNN), a gradient-trained controller whose topology evolves online through a local Structural Plasticity Module (SPM). The SPM monitors edge-wise weight update statistics over short temporal windows and uses these local signals to trigger neuron insertion and pruning, while synaptic weights are optimized by a standard gradient-based optimizer. This allows network capacity to be adjusted during learning without manual architectural tuning.

SMGrNN is evaluated on control benchmarks via policy distillation. Compared with multilayer perceptron baselines, it achieves similar or higher returns, lower variance, and task-appropriate network sizes. Ablation studies with growth disabled and growth-only variants isolate the role of structural plasticity, showing that adaptive growth improves reward stability while pruning prevents uncontrolled expansion and supports compact network formation. These results establish the independent value of local structural plasticity within gradient-trained networks and motivate future investigation of whether similar structural rules can be extended to more local or spike-based learning settings.
\end{abstract}


\section{Introduction}
\label{intro}
\subsection{Research Background and Significance}
\IEEEPARstart{D}{eep} neural networks have become standard tools for learning control policies, yet most practical architectures are still designed with \emph{fixed capacity} and trained with \emph{task-specific supervision}. In typical deep reinforcement learning pipelines, controllers are implemented as multilayer perceptrons with a pre-specified number of neurons and optimized end-to-end by backpropagation under carefully engineered reward signals. This combination of fixed capacity and reliance on global error information leads to two interrelated limitations.

First, traditional paradigms rely heavily on external reward signals or manually labeled data, which are costly to obtain and difficult to generalize across tasks. Deep learning typically depends on backpropagation, a non-local mechanism that requires propagating global error gradients through all layers \cite{konishi2023biologically,bengio2015towards}; even in reinforcement learning, controllers require carefully designed reward functions \cite{stegmaier2023biologically,doya2007reinforcement}. Human-in-the-loop approaches can mitigate some of these issues but still presuppose external guidance. Biological systems, in contrast, adapt through local learning rules that update synapses based on locally available signals such as pre-post activity correlations \cite{hebb2005organization}. Recent work has shown that Hebbian-style rules can train deep networks or shape useful representations without explicit global error propagation \cite{journe2022hebbian,talloen2021pytorch}. Taken together, these observations highlight \emph{local interaction rules} as a core organizing principle: synaptic strengths and connectivity motifs are shaped by correlations in neural activity rather than by global error signals. This perspective motivates control architectures whose connectivity is governed by local structural signals, while synaptic weights are still optimized by conventional gradient-based methods.

Second, fixed architectures constrain long-term adaptation and can exacerbate catastrophic forgetting in non-stationary environments \cite{parisi2019continual,van2024continual,kirkpatrick2017overcoming,wang2024comprehensive}. Biological networks exhibit structural plasticity, reorganizing neurons and synapses throughout life \cite{marzola2023exploring}, while most artificial networks, in contrast, remain static and brittle in the face of changing conditions \cite{markram2011history}. The need to predefine network capacity often leads to under-utilized resources when overparameterized, or loss of performance and interference once capacity is saturated \cite{sodhani2020toward,chen2018lifelong,huang2021understanding}. Existing continual learning methods can expand capacity \cite{rostami2019complementary,rolnick2019experience}, but expansions are typically manual or governed by hand-designed rules rather than emerging from the dynamics of learning itself. These limitations motivate neural controllers in which architectural growth and pruning are driven by internal edge-wise weight-update statistics, allowing capacity to self-adjust to task demands.

The Self-Motivated Growing Neural Network (SMGrNN) is introduced as a gradient-trained neural controller whose topology evolves online through a structural plasticity module (SPM). Here, ``self-motivated'' means that structural changes are driven by internally available edge-local statistics rather than by task-specific growth schedules or manually specified architecture updates. In the current formulation, synaptic weights are optimized by standard backpropagation, while the SPM regulates architectural capacity during training through local structural decisions. The purpose of this design is not to claim fully local learning, but to isolate and evaluate the independent role of local structural plasticity within a standard gradient-trained setting. This separation allows the present study to ask whether local structural adaptation alone can improve stability and match network capacity to task demands within a single task.

\subsection{Research Objectives}
This study aims to examine the role of local structural plasticity in gradient-trained neural controllers, with a focus on within-task capacity adaptation rather than continual learning across task sequences. The specific objectives are threefold. First, to formulate a Structural Plasticity Module (SPM) that uses local edge-level statistics to decide when and where to grow or prune neurons and connections. Second, to instantiate the SPM within SMGrNN and systematically evaluate its effect on policy distillation for control benchmarks, including comparisons with fixed-capacity multilayer perceptrons and ablations with growth disabled or restricted to growth-only settings. Third, to clarify the separation between structural plasticity and synaptic optimization in the current framework, thereby identifying what can already be established in a gradient-trained setting and what remains for future local-learning extensions.

By addressing these objectives, SMGrNN is intended to clarify the independent contribution of local structural plasticity to learning stability, compactness, and task-matched capacity adaptation within gradient-trained networks.

\section{Related Works}

This section reviews work on structural adaptation and neural architecture growth, as well as biologically motivated local learning algorithms, and identifies the gaps that motivate the proposed SMGrNN framework.

\subsection{Literature Review}

\subsubsection{Structural Adaptation in Continual Learning Models}

Over the past decade, dynamic neural architectures have been explored in continual and sequential learning settings as a way to allocate new capacity rather than overwrite existing parameters. Progressive Neural Networks (PNN) \cite{rusu2016progressive} freeze old parameters and add new columns per task, avoiding forgetting but requiring task boundaries and growing unboundedly. Later methods improved efficiency: Dynamically Expandable Networks (DEN) \cite{yoon2017lifelong} selectively add and prune neurons based on utility, while Neurogenesis Deep Learning \cite{draelos2017neurogenesis} inserts neurons for novel classes inspired by adult neurogenesis.

More recent work refines expansion without explicit task boundaries. The Self-Controlled Dynamic Expansion Model (SCDEM) \cite{wu2025self} spawns lightweight task-specific experts with collaborative optimization across backbones, while the Self-Evolved Dynamic Expansion Model (SEDEM) \cite{ye2023self} instantiates new experts when novelty exceeds a threshold, reusing prior features via dynamic masks. Both highlight data-driven criteria for when and how much to expand.

Other approaches focus on modularity and compression. The Modular Dynamic Neural Network (MDNN) \cite{turner2021modular} grows a tree of sub-networks to localize changes and reduce interference. Yet unbounded growth remains a challenge: DEN, SCDEM, and SEDEM mitigate it through selective expansion and pruning, while PNN and naive modular networks grow linearly. Compression-based methods such as Progress \& Compress \cite{schwarz2018progress} and Bayesian non-parametric models \cite{lee2020neural} aim for bounded or sublinear scaling.

Finally, in neuromorphic systems, DSD-SNN \cite{han2023enhancing} demonstrates that growth and pruning of spiking neurons also improve performance in continual learning settings, underscoring architectural plasticity as a principle bridging machine and biological networks.

\subsubsection{Growing and Self-Assembling Neural Networks}

Early evidence that structure can be learned comes from neuro-evolution: NeuroEvolution of Augmenting Topologies (NEAT) \cite{stanley2002evolving} evolves both weights and topology from minimal seeds via mutations. While effective, such methods are typically offline and evaluation-heavy, limiting practicality for online adaptation.

More recent, biologically motivated work treats growth as a decentralized developmental process. HyperNCA \cite{najarro2022hypernca} and NDP \cite{najarro2023towards} use local rules to self-assemble functional architectures; however, growth usually occurs in an initial phase and task learning proceeds with a fixed structure thereafter.

LNDP \cite{plantec2024evolving} extends NDPs to ongoing plasticity throughout the agent's lifetime, adding and removing synaptic connections based on local neuronal activity and a global reinforcement signal. It achieves dynamic structural adaptation for sequential control and outperforms fixed-topology baselines, but relies on meta-optimized growth and pruning policies and performs only limited pruning, which may complicate interpretation and raise concerns about long-horizon scalability.

Relatedly, Growing Neural Cellular Automata \cite{mordvintsev2020growing} shows that complex global patterns can emerge from learned local rules, suggesting that open-ended architectural growth is feasible provided growth remains goal-directed and computationally tractable.

\subsubsection{Hebbian Learning in Deep Networks}

Backpropagation depends on global error transport, whereas biological systems rely on local updates. Recent work shows that local rules can scale to deep models: SoftHebb \cite{journe2022hebbian} trains networks with layer-wise Hebbian objectives and achieves competitive classification, while the Forward-Forward (FF) algorithm \cite{hinton2022forward} uses positive and negative passes with local ``goodness'' signals. Theoretically, Hebbian self-organization with synaptic turnover yields heavy-tailed connectivity like that observed in brains \cite{lynn2024heavy}, suggesting that local plasticity not only learns representations but also shapes network structure. These results indicate that synaptic learning rules need not rely exclusively on global error signals and motivate structural plasticity mechanisms that are likewise local and compatible with Hebbian or spike-based updates, even when gradient-based optimization is used in practice.

\begin{figure*}
    \centering
    \includegraphics[width=0.8\linewidth]{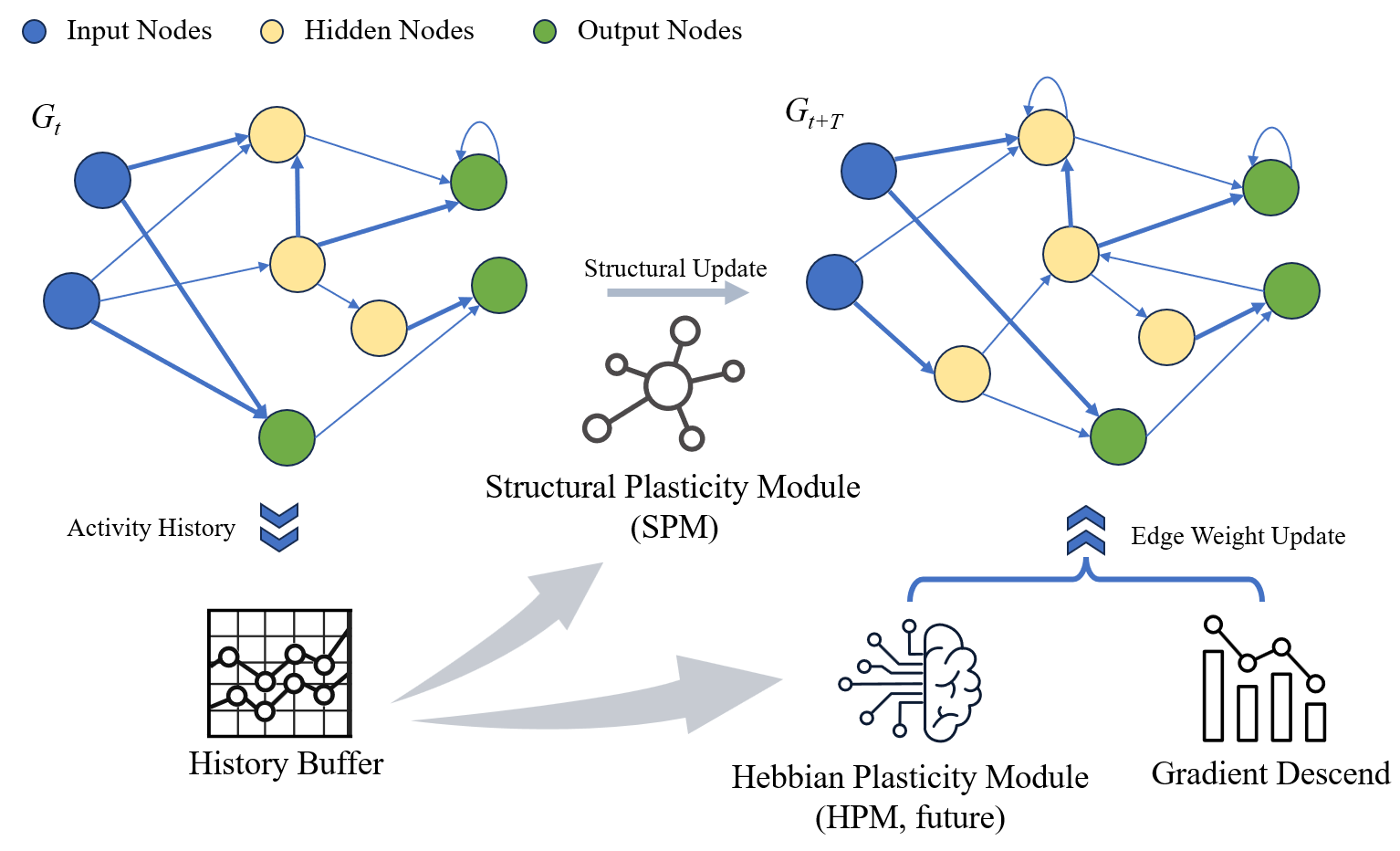}
    \caption{Overview of the SMGrNN.
    The current graph $G_t$ receives local update statistics from history buffers,
    which are processed by the SPM to produce
    an updated graph (denoted $G^*$ in the figure). Synaptic weights are
    updated by a gradient-based optimizer on the right. The figure emphasizes the separation between local structural adaptation and gradient-based synaptic optimization in the current work. Any future local synaptic plasticity module is outside the scope of the experiments reported here.}
    \label{fig:smgrnn_overview}
\end{figure*}

\subsection{Research Gap and Proposed Contribution}

In summary, the literature confirms two points: (1) allowing a network's architecture to expand or reconfigure over time is an effective way to adapt capacity and reduce interference across tasks or data regimes \cite{rusu2016progressive,yoon2017lifelong,turner2021modular,han2023enhancing,plantec2024evolving}; (2) learning algorithms with reduced dependence on global error signals---such as Hebbian or layer-local objectives---are viable and biologically motivated \cite{journe2022hebbian,hinton2022forward,lynn2024heavy}. However, existing dynamic architecture methods (e.g., PNN, DEN, SCDEM, SEDEM) still rely on backpropagation within each module and typically decide when and where to expand based on task boundaries, global performance criteria, or hand-designed heuristics rather than purely local edge-wise update statistics. Conversely, purely Hebbian or feedback-free approaches generally assume fixed architectures and rarely analyze how structural adaptation interacts with policy learning, particularly in control settings. The present work therefore focuses on within-task capacity adaptation, rather than on mitigating catastrophic forgetting across task sequences; explicit continual learning is left to future work.

The proposed SMGrNN framework addresses this gap by introducing a local SPM that grows and prunes neurons based solely on short-term edge-local statistics of weight change. This module is instantiated within a gradient-trained policy network and evaluated on control benchmarks to quantify its impact on reward stability, variance, compactness, and discovered network size relative to fixed-capacity multilayer perceptrons and ablated variants. The central contribution of the present work is therefore not a fully local learning system, but a demonstration that local structural plasticity can already provide measurable benefits within a standard gradient-trained framework. In this sense, the paper focuses on within-task adaptive capacity control as an independent result, while future combinations with more local synaptic learning rules are left for subsequent work.

\section{Methodology}
\label{sec:method}

The SMGrNN is a directed graph
$G_t = (V_t, E_t)$ whose nodes are neurons and whose directed, weighted edges are synapses.
Unlike fixed multilayer perceptrons, $G_t$ changes during training: edges are added or removed
and hidden nodes are inserted or deleted by a SPM, while synaptic
weights are updated by a standard gradient-based optimizer.
Short, rolling history buffers store recent edge-wise weight updates; the resulting local temporal statistics drive structural decisions in the SPM.
This section formalizes the control setting, the SMGrNN architecture, the SPM rules, and the
training procedure used in the experiments.
Fig.~\ref{fig:smgrnn_overview} summarizes the separation between local structural adaptation in the SPM and gradient-based synaptic optimization in the current implementation.

\subsection{Control Setting and Policy Representation}

Consider a Markov decision process with state $s_t \in \mathcal{S}$ and action
$a_t \in \mathcal{A}$.
A fixed expert policy $\pi_E(a \mid s)$ is implemented by a pre-trained multilayer perceptron (MLP).
The objective is to learn a student policy $\pi_\theta(a \mid s)$ parameterized by SMGrNN
that imitates the expert while autonomously adjusting its own capacity.

At each interaction step, the environment state $s_t$ is written to the input
nodes of the current graph $G_t$. The graph then performs $K$ synchronous
message-passing iterations, producing an output action
$\hat{a}_t = \pi_\theta(s_t)$ from the output nodes after the final iteration.
A supervised loss $L_t(\theta)$ penalizes the deviation between the student
and expert actions, for example as a mean-squared error for continuous
actions or a cross-entropy for discrete action distributions. Parameters
$\theta$ (edge weights) are updated by a gradient-based optimizer through
the computation graph induced by these $K$ explicit propagation steps,
while the SPM monitors local edge-wise weight-update statistics to decide
when and where to grow or prune the architecture.

\subsection{Self-Motivated Growing Neural Network}

SMGrNN is instantiated as a directed graph whose nodes are functionally
grouped into input, hidden, and output sets. Let $V_t$ denote the node set
at step $t$, $E_t \subseteq V_t \times V_t$ the directed edge set, and
$w_k^{(t)}$ the scalar weight on edge $k \equiv (i \rightarrow j) \in E_t$.
Input and output nodes are fixed, whereas hidden nodes may be inserted or
removed during training.

The graph is not restricted to a layered multilayer-perceptron topology.
In the current implementation, skip connections and feedback edges may
emerge as structural plasticity edits accumulate. Accordingly, forward
propagation is not defined by a single topological pass. Instead, SMGrNN
is treated as a graph-state system whose node states are updated through
a fixed number of synchronous message-passing iterations.

Let $\mathbf{x}_t^{(0)} \in \mathbb{R}^{|V_t| \times F}$ denote the current
node-state matrix at training step $t$, after clamping the input-node states
to the external observation. For iteration $r = 0, \dots, K-1$, the network
computes
\[
\mathbf{m}_t^{(r+1)} = \mathrm{Propagate}\!\left(E_t, \mathbf{x}_t^{(r)}, \mathbf{w}_t\right),
\]
and then updates non-input nodes synchronously as
\[
\mathbf{x}_{t,i}^{(r+1)} =
\begin{cases}
\mathbf{x}_{t,i}^{(r)}, & i \in V_{\mathrm{in}},\\[4pt]
\phi_{\mathrm{out}}\!\left(\mathbf{m}_{t,i}^{(r+1)}\right), & i \in V_{\mathrm{out}},\\[4pt]
\phi\!\left(\mathbf{m}_{t,i}^{(r+1)}\right), & i \in V_{\mathrm{hid}},
\end{cases}
\]
where $\phi$ is the hidden-node activation function and $\phi_{\mathrm{out}}$
is the output-node activation function. In this way, cycles are handled
implicitly through iterative state updates rather than through an acyclic
evaluation order.

After $K$ iterations, the output-node states are taken as the student-policy
output. The resulting node states are written back into the graph as the
internal state for the next call, but they are detached from the autograd
graph after each forward pass. Therefore, gradients are propagated through
the $K$ explicitly unrolled message-passing iterations within one forward
call, but not across multiple forward calls over time.

\subsection{Structural Plasticity Module (SPM)}
\label{sec:spm}
\subsubsection{Edge-Driven Growth}

For each edge $k \equiv (i \rightarrow j)$ with weight $w_k$, let
$\Delta w_{k,t}$ denote its instantaneous update at step $t$ as produced by
the gradient-based optimizer. Over a sliding window of $T$ steps, the SPM
maintains the sample mean and variance
\begin{equation}
\begin{aligned}
\mu_k^\Delta
&=\frac{1}{T}\sum_{\tau=t-T+1}^{t}\Delta w_{k,\tau},\\
(\sigma_k^\Delta)^2
&=\frac{1}{T}\sum_{\tau=t-T+1}^{t}
\left(\Delta w_{k,\tau}-\mu_k^\Delta\right)^2 .
\end{aligned}
\label{eq:update_stats}
\end{equation}

To detect edges whose update direction remains unresolved over the observation
window, define the local fluctuation interval
\begin{equation}
I_k^\Delta=
\left[
\mu_k^\Delta-\lambda \sigma_k^\Delta,\,
\mu_k^\Delta+\lambda \sigma_k^\Delta
\right],
\label{eq:fluctuation_interval}
\end{equation}
where $\lambda > 0$ controls the width of the interval. The condition
$0 \in I_k^\Delta$ indicates that the update direction fluctuates around zero
rather than settling on a clear positive or negative trend.

To distinguish meaningful fluctuation from negligible optimizer noise, the
standard deviation is further normalized by a regularized local update scale:
\begin{equation}
\tilde{w}_k = |\mu_k^\Delta| + \epsilon,
\qquad
\tilde{\sigma}_k^\Delta = \frac{\sigma_k^\Delta}{\tilde{w}_k},
\label{eq:normalized_fluctuation}
\end{equation}
where $\epsilon > 0$ is a small constant that keeps the denominator
well-defined when the mean update is close to zero. An edge is then flagged
as unstable if
\begin{equation}
0 \in I_k^\Delta
\quad \land \quad
(\tilde{\sigma}_k^\Delta)^2 > \theta_{\mathrm{var}},
\label{eq:unstable_edge}
\end{equation}
where $\theta_{\mathrm{var}} > 0$ is a threshold on the normalized fluctuation
magnitude. The first condition identifies edges whose update direction remains
unresolved, while the second requires that this fluctuation be sufficiently
large relative to the local update scale.

\begin{figure}[t]
    \centering
    \includegraphics[width=0.9\linewidth]{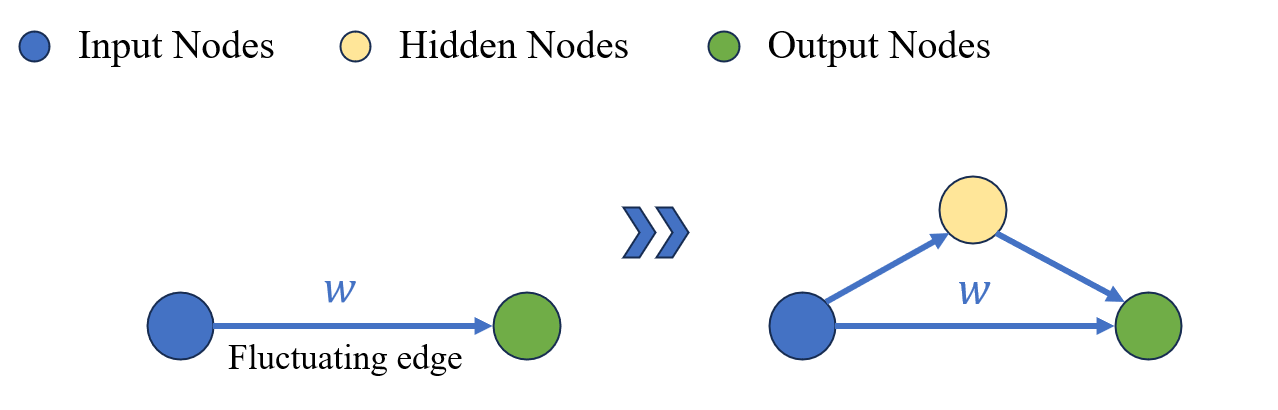}
    \caption{Relay-node insertion on an unstable edge in a minimal network.
    Left: a single input--output connection whose update direction remains
    unresolved over the observation window and whose normalized fluctuation
    satisfies Eq.~(\ref{eq:unstable_edge}).
    Right: when the edge is flagged as unstable, the SPM inserts a new relay
    node and a two-step pathway, while retaining the original edge. The example
    shows a one-input, one-output network for clarity, but the same mechanism
    can be applied to any edge in a larger SMGrNN graph that satisfies
    Eq.~(\ref{eq:unstable_edge}).}
    \label{fig:relay_insertion}
\end{figure}

When Eq.~(\ref{eq:unstable_edge}) holds, the SPM inserts a new relay node
$h_{\mathrm{new}}$ and adds a parallel two-step pathway
$i \rightarrow h_{\mathrm{new}} \rightarrow j$ alongside the original edge.
The new weights are initialized with small magnitude $|w_{\mathrm{init}}| \ll 1$
so that the added path does not disrupt behavior at insertion time. The
original edge $i \rightarrow j$ is retained, allowing both direct and
relay-mediated signal propagation. In this way, SMGrNN locally increases
representational capacity where the update statistics indicate persistent
structural uncertainty. As illustrated in Fig.~\ref{fig:relay_insertion}, the relay node is introduced
on a fluctuating edge in a minimal one-input/one-output example, but the same
operation can be applied to any edge in $G_t$ that satisfies
Eq.~(\ref{eq:unstable_edge}).

The relay insertion operator is chosen as a simple and conservative local
capacity-increase mechanism. Because the instability signal is detected on a
specific edge, the structural edit is also applied locally around that same
edge by adding a parallel two-step pathway
\(i \rightarrow h_{\mathrm{new}} \rightarrow j\) while retaining the original
edge \(i \rightarrow j\). This design has two practical motivations. First,
it is a minimal local edit aligned with the edge-local trigger itself, rather
than a more global rewiring operation. Second, by preserving the original
edge and initializing the new pathway with small weights, the model can
increase representational capacity without abruptly disrupting the existing
signal path at insertion time.

\subsubsection{Random Exploratory Growth}
Edge-driven growth reacts only to edges that already exist.
To explore alternative connectivity patterns and avoid getting trapped in suboptimal local
structures, the SPM also performs occasional random exploratory growth.

Let $N_t = \lvert V_t \rvert$, where $V_t$ and $E_t$ denote the node and edge sets at step $t$.
The SPM first samples a Bernoulli trigger
$B_t \sim \mathrm{Bernoulli}(p_{\mathrm{rand}})$, and forces $B_t = 1$ if no
criterion-driven growth candidates exist. Here, criterion-driven growth candidates
refer to edges that satisfy the edge-driven instability criterion in Eq.~(4).
When $B_t = 1$, it proposes
\begin{equation}
m_t = \max\bigl(\lfloor \rho_{\mathrm{rand}} N_t \rfloor, 1\bigr)
\label{eq:exploratory-batch}
\end{equation}
new edges by drawing pairs without replacement from a candidate pool $S_t$, defined as
\begin{equation}
S_t = \{(u, v) \in V_t \times V_t : u \neq v,\ (u, v) \notin E_t\}.
\label{eq:exploratory-pool}
\end{equation}
The proposed edges $E_t^+ = \{(u_\ell \to v_\ell)\}_{\ell=1}^{m_t}$ are then sampled from
$S_t$ and added with weights initialized to $w_{\mathrm{init}}$.
The hyperparameters $p_{\mathrm{rand}} \in (0,1)$ and $\rho_{\mathrm{rand}} \in (0,1)$
control the frequency and batch size of exploratory growth.
In expectation,
$\mathbb{E}[m_t B_t] \approx \rho_{\mathrm{rand}} N_t p_{\mathrm{rand}}$
when criterion-driven growth is available.

\subsubsection{Pruning and Orphan Removal}

To prevent unbounded expansion and remove ineffective connections, the SPM performs
weight-based pruning.
For each edge $k$, the module defines a candidate removal set
\begin{equation}
W_t = \Bigl\{
k : \lvert w^{(t)}_k \rvert \le \tau_w \ \wedge\ 
\lvert \mu^{\Delta}_k \rvert \le \tau_{\Delta}
\Bigr\},
\label{eq:pruning-set}
\end{equation}
where $\tau_w > 0$ suppresses weights with small magnitude and $\tau_{\Delta} > 0$
removes edges whose updates have effectively stalled.
From $W_t$, only a fraction $\eta_{\text{prune}} \in (0,1]$ is actually deleted at each
pruning step, chosen uniformly at random to avoid abrupt structural changes.

Pruning edges may leave hidden nodes with zero in-degree or zero out-degree.
Let $\deg^{\text{in}}_t(i)$ and $\deg^{\text{out}}_t(i)$ denote the in- and out-degrees
of node $i$ at step $t$, and let $H \subseteq V_t$ be the set of hidden nodes.
The SPM removes orphan nodes via
\begin{equation}
\label{eq:orphan-set}
U_t = \Bigl\{
i \in H : \deg^{\text{in}}_t(i) = 0 \ \vee\ 
\deg^{\text{out}}_t(i) = 0
\Bigr\},
\end{equation}
deleting $U_t$ and all incident edges.
This maintains a compact active subgraph and ensures that hidden units contribute
either to processing inputs or to driving outputs.

\subsubsection{Schedule and Net Growth Metrics}

Growth and pruning are interleaved according to a simple schedule.
At each step $t$, the SPM selects
\begin{equation}
\Sigma(t) =
\begin{cases}
\text{prune}, & t \equiv 0 \ (\text{mod } s), \\
\text{grow},  & \text{otherwise},
\end{cases}
\end{equation}
where $s \in \mathbb{N}$ is a pruning period.
During growth steps the edge-driven and exploratory mechanisms are applied; during pruning
steps Eqs.~\eqref{eq:pruning-set}--\eqref{eq:orphan-set} are used to trim weak or inactive structure.

Let $\Delta N_t = \lvert V_{t+1} \rvert - \lvert V_t \rvert$ and
$\Delta M_t = \lvert E_{t+1} \rvert - \lvert E_t \rvert$ denote the per-step change
in node and edge counts.
Net growth over a training run is reported as
\begin{equation}
\text{NetGrowth}_{\text{nodes}} = \sum_t \Delta N_t,
\qquad
\text{NetGrowth}_{\text{edges}} = \sum_t \Delta M_t,
\end{equation}
which summarizes the total structural expansion induced by the SPM.

\subsection{History Buffers and Local Statistics}

The SPM relies on short-term temporal statistics rather than instantaneous values.
To this end, SMGrNN maintains rolling buffers for edge-wise weight updates.
For each edge $k$, the buffer stores recent update increments $\Delta w_{k,t}$
over the last $T$ steps.
These buffers are updated online after each parameter update and reset whenever
a structural edit modifies the edge indexing.
Local statistics such as $\mu_k^{\Delta}$ and $(\sigma_k^{\Delta})^2$ are computed
directly from the buffers, providing the SPM with an intrinsic view of which
connections are stable, unstable, or inactive, without accessing any global loss information.

\subsection{Training Procedure}
Training proceeds as online policy distillation in Gym control environments.
At each episode, the current SMGrNN policy interacts with the environment,
generating a trajectory $(s_t, a_t)$. For each visited state $s_t$, the expert
MLP produces a reference action $a_t^E = \pi_E(s_t)$, and SMGrNN produces
$\hat{a}_t = \pi_\theta(s_t)$ after $K$ synchronous message-passing iterations
of the current graph $G_t$. A supervised loss $L_t(\theta)$ is computed
between $a_t^E$ and $\hat{a}_t$ and used to update the trainable weights by
backpropagation through the $K$ explicitly unrolled propagation steps.

Importantly, the node states produced at the end of a forward call are stored
as the graph's internal state for the next call, but are detached from the
autograd graph before reuse. Thus, the current implementation performs
backpropagation only through the finite set of propagation iterations within
each forward call, rather than full backpropagation through time across
multiple interaction steps.

The SPM is invoked according to the structural schedule $\Sigma(t)$ to adjust
the graph based on the latest local statistics from the history buffers. In the
experiments, the same training procedure is applied across tasks, with
differences only in environment dynamics and expert policies.
\subsection{Implications for Local Synaptic Plasticity}

The current experiments do not validate SMGrNN under Hebbian, spike-timing-dependent, or other fully local synaptic learning rules. In this work, synaptic weights are optimized by backpropagation, and the empirical contribution is limited to showing that local structural plasticity can be studied independently within a gradient-trained network.

The separation between structural adaptation and synaptic optimization nevertheless provides a useful conceptual boundary for future work. Because the SPM operates on locally available edge-level statistics, one may ask whether similar structural rules could be reformulated for settings with more local synaptic updates. However, such an extension is not guaranteed. In particular, the meaning of weight-change statistics may differ substantially between gradient-based learning and Hebbian or spike-based plasticity; in spiking systems, variability in local updates may also reflect encoding stochasticity rather than unresolved representation. Accordingly, the present paper does not claim compatibility in a validated sense, but only identifies a possible direction for future investigation.

\section{Experiments}

\subsection{Experimental Setup}

\subsubsection{Tasks and expert policies}
Three classic control environments from the Gym suite are considered for
the main policy-distillation experiments: CartPole-v1, Acrobot-v1, and
LunarLander-v3. In addition, a simple XOR classification task is included in
the task-difficulty scaling analysis as a minimal non-control reference.

For the three control environments, a pre-trained multilayer perceptron
(MLP) serves as a fixed expert policy $\pi_E(a \mid s)$ and provides target
actions for policy distillation; the expert never acts on the environment
during student training. In the current implementation, each expert is a
single-hidden-layer ReLU MLP with 2048 hidden units and one output per
discrete action. Expert models are trained separately using a DQN-style
procedure with Adam optimization (learning rate $10^{-3}$), discount factor
$\gamma=0.99$, $\epsilon$-greedy exploration, a replay buffer of size
10{,}000, batch size 64, and parameter updates every 4 environment steps.
The best expert checkpoint is then frozen and used for subsequent student
distillation. During distillation, the expert's greedy action is obtained by
$\arg\max$ over its action scores and converted into the target action label
for the student. The student policy $\pi_\theta$ is parameterized either by
SMGrNN or by a static MLP baseline.

The XOR task is not used for expert distillation. Instead, it is treated as a
small supervised problem used only in the task-difficulty scaling analysis,
where the objective is to examine how the learned network size changes
from a trivial classification task to progressively harder control tasks.

Unless noted otherwise, each control-task condition is trained for 2000
episodes at an initial connection density of 0.8 and is repeated with 10
random seeds. During training, the following quantities are recorded:
\begin{enumerate}
    \item episodic return (mean with across-run variability);
    \item network size trajectory (number of hidden nodes and edges);
    \item convergence episode to a preset reward threshold for each environment; and
    \item late-training return, defined as the average reward over the final quarter of episodes.
\end{enumerate}

\subsubsection{SMGrNN configuration}
\label{sec:exp_smgrnn_config}

SMGrNN is initialized with fixed input and output nodes matching the
state and action dimensions of each environment. Unless otherwise stated,
the hidden-node count is initialized to zero, so the initial architecture
contains only direct connections between inputs and outputs. Edges are
instantiated with a target connection density of $0.8$ over allowed node
pairs, and all neurons use a $\tanh$ nonlinearity. Weights are drawn from
a zero-mean Gaussian with small variance.

In the control experiments reported here, each forward call uses the
conservative setting \(K=1\) synchronous message-passing iteration. The
node-state feature dimension is \(F=1\), so each node maintains a scalar
internal state. The regularization constant in Eq.~(3) is set to
\(\epsilon=10^{-8}\) to avoid division by zero when the mean edge update
is close to zero. With \(K=1\), feedback edges and cycles are not evaluated
by solving a fixed point within a single forward call. Instead, their effect
enters through the carried node states across successive calls. At the
beginning of each episode, hidden and output node states are reset to zero.
Within an episode, the node states produced at the end of a forward call are
stored as the graph's internal state for the next call, but are detached from
the autograd graph before reuse. Therefore, gradients are propagated only
through the single synchronous message-passing step within each forward
call, not through the full sequence of environment interactions.

For the XOR experiment, we use \(K=3\). This is because XOR is evaluated
as a single-step forward computation with node states reset at the start, so
cross-call state carry-over is unavailable and limited within-call iterative
propagation is needed instead. The present paper therefore covers two
limited regimes: \(K=1\) for the control tasks, where recurrent connectivity
is mediated mainly through cross-call state, and \(K=3\) for XOR, where a
small amount of within-call propagation is used in the absence of such state
carry-over. This is a deliberate engineering choice for the present study,
rather than a claim to solve unrestricted recurrent credit assignment on
arbitrary graphs.

Unless otherwise stated, all gradient-trained student models in the main
experiments use the same optimizer configuration, namely Adam with
learning rate \(0.01\), so that structural comparisons are not confounded
by differences in optimizer choice or learning-rate scale across conditions. A small CartPole-v1 sensitivity study on optimizer and learning-rate choice
is reported in Appendix~\ref{app:optimizer_lr_sensitivity}. In brief, growth
and pruning remain active across the tested optimizers and learning rates,
while optimizer choice and learning-rate scale mainly affect the magnitude
and stability of structural adaptation.

\subsubsection{SPM hyperparameter selection}
\label{sec:exp_hparam_selection}
The SPM uses a temporal window of length $T$ to compute local statistics
of edge-wise weight updates. Edge-driven growth is governed by the
instability criterion in Eq.~(\ref{eq:unstable_edge}), which depends on
$\lambda$ and $\theta_{\mathrm{var}}$. Random exploratory growth is
controlled by $p_{\mathrm{rand}}$ and $\rho_{\mathrm{rand}}$, while pruning
is controlled by $\tau_w$, $\tau_\Delta$, $\eta_{\mathrm{prune}}$, and $s$.
The numerical values used in the main experiments are listed in
Table~\ref{tab:spm_hyperparams}.

To make hyperparameter selection reproducible, we used a two-stage
protocol. First, a $3^4$ grid search on CartPole-v1 tuned the four primary
SPM parameters $(T,\lambda,\theta_{\mathrm{var}},p_{\mathrm{rand}})$.
The remaining SPM hyperparameters were held fixed at conservative defaults
because they mainly regulate secondary aspects of structural editing, such
as pruning frequency, pruning aggressiveness, and random-growth batch size.

Second, one-factor-at-a-time sensitivity analyses were performed on
CartPole-v1 and Acrobot-v1 around the tuned region. These experiments
showed that $T$ had the clearest task dependence, whereas
$\theta_{\mathrm{var}}$ mainly controlled the compactness of the final
network. The final shared defaults used in the main experiments were
therefore chosen as a practical cross-task compromise rather than as the
single-task optimum found on CartPole-v1. Full search ranges, selection
criteria, and additional sensitivity results are provided in
Appendix~\ref{app:hparam_selection}.

\subsubsection{Baseline configurations}
\label{sec:exp_baseline_config}

We compare SMGrNN against both fixed-topology and dynamic-topology
baselines. The fixed-topology baselines include Static-SMGrNN and
parameter-matched MLPs. Static-SMGrNN shares the same initialization and
training hyperparameters as SMGrNN but disables the SPM, so no structural
edits occur during training. The MLP baselines are conventional
single-hidden-layer networks constructed with environment-specific hidden
widths corresponding approximately to fixed parameter budgets relative to
the final SMGrNN size.

To strengthen the comparison with existing dynamic architecture methods,
we additionally implement two representative dynamic-topology baselines:
NEAT and a Neural Developmental Program (NDP) baseline. NEAT represents
an evolutionary approach that jointly optimizes network topology and
weights, whereas NDP represents a developmental growth process that
constructs network structure through repeated local development steps.
Both baselines are adapted to the same control benchmark tasks and evaluated
on CartPole-v1, Acrobot-v1, and LunarLander-v3 using 10 random seeds.

For budget control, all methods are allocated the same total number of
environment interaction rollouts per seed. Gradient-trained student models
are trained for 2000 online episodes, whereas NEAT and NDP are evaluated
with a population size of 50 for 40 generations, corresponding to
$50 \times 40 = 2000$ candidate-policy evaluation rollouts per seed. For
NDP, each candidate policy is generated with 6 developmental steps before
evaluation. Because NEAT and NDP are evolutionary or developmental
methods rather than gradient-trained online distillation models, their
evaluation curves are reported differently from those of SMGrNN and the
MLP baselines.
For SMGrNN and MLPs, curves represent
online reward trajectories during gradient-based student training. For NEAT
and NDP, cross markers denote the performance of the best individual selected
in each evolutionary generation, and the accompanying smoothed curves are
used only to visualize the overall search/evaluation trend. Therefore, these
dynamic-topology baselines should be interpreted as adapted comparisons
under the same benchmark tasks, rather than as full reproductions of their
original experimental protocols or as optimization trajectories directly
equivalent to online distillation.

\subsection{Comparison with Fixed and Dynamic Topology Baselines}
\label{sec:baseline_comparison}

To evaluate whether the benefits of SMGrNN arise from adaptive structural
plasticity rather than from parameter count or dynamic topology alone, we
compare it with both fixed-topology and dynamic-topology baselines. The
fixed-topology baselines include parameter-matched MLPs with approximately
$100\%$ and $200\%$ of the final SMGrNN parameter budget. The
dynamic-topology baselines include NEAT and NDP, which represent
evolutionary topology optimization and developmental network growth,
respectively. All methods are evaluated on CartPole-v1, Acrobot-v1, and LunarLander-v3
using 10 random seeds under the same 2000 episode-equivalent rollout budget
per seed.

\begin{figure*}[t]
    \centering

    \begin{subfigure}[b]{0.32\textwidth}
        \centering
        \includegraphics[width=\linewidth]{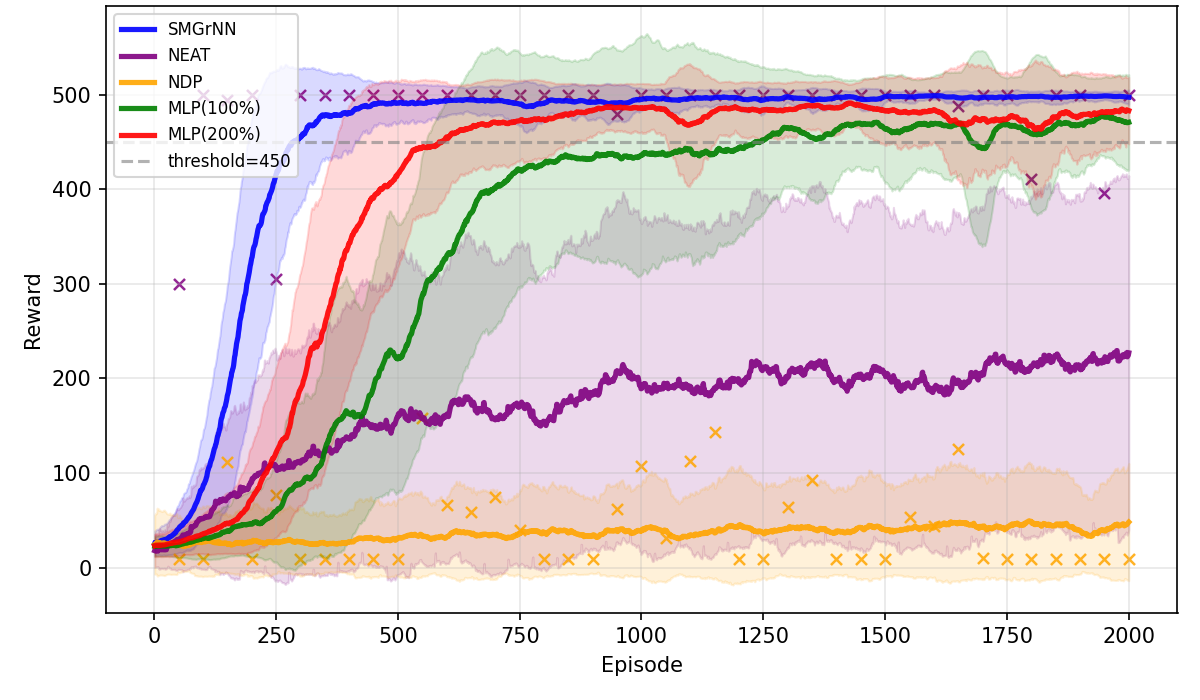}
        \caption{CartPole-v1 reward}
        \label{fig:baseline_reward_cartpole}
    \end{subfigure}
    \hfill
    \begin{subfigure}[b]{0.32\textwidth}
        \centering
        \includegraphics[width=\linewidth]{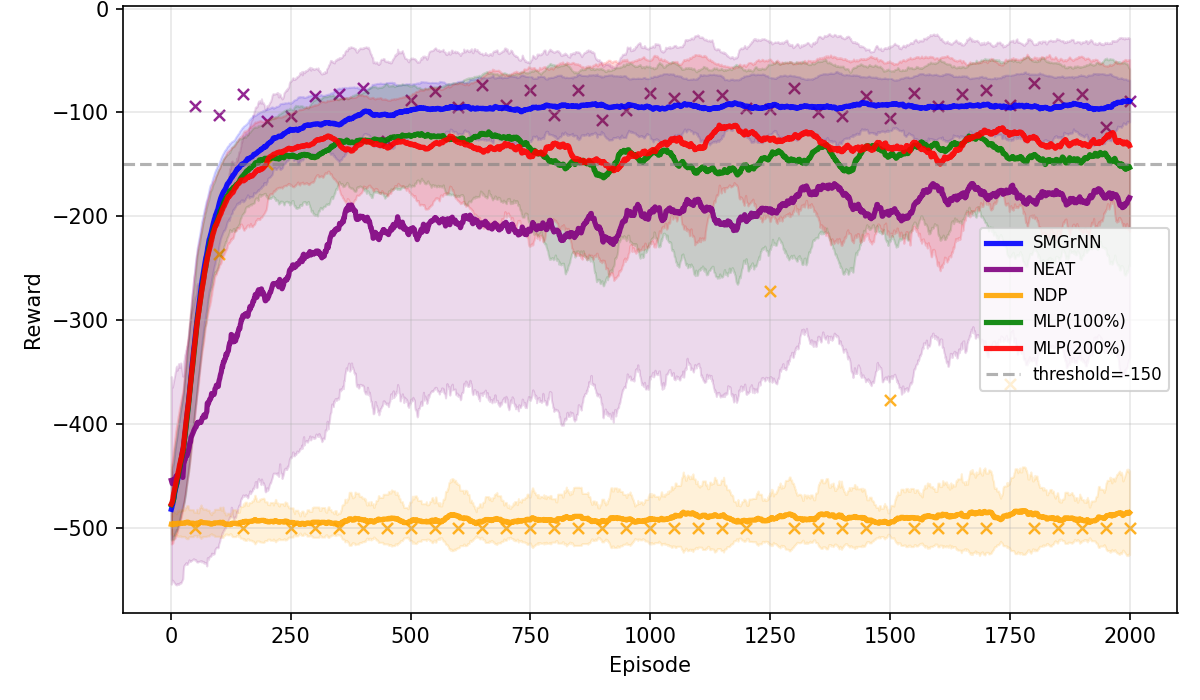}
        \caption{Acrobot-v1 reward}
        \label{fig:baseline_reward_acrobot}
    \end{subfigure}
    \hfill
    \begin{subfigure}[b]{0.32\textwidth}
        \centering
        \includegraphics[width=\linewidth]{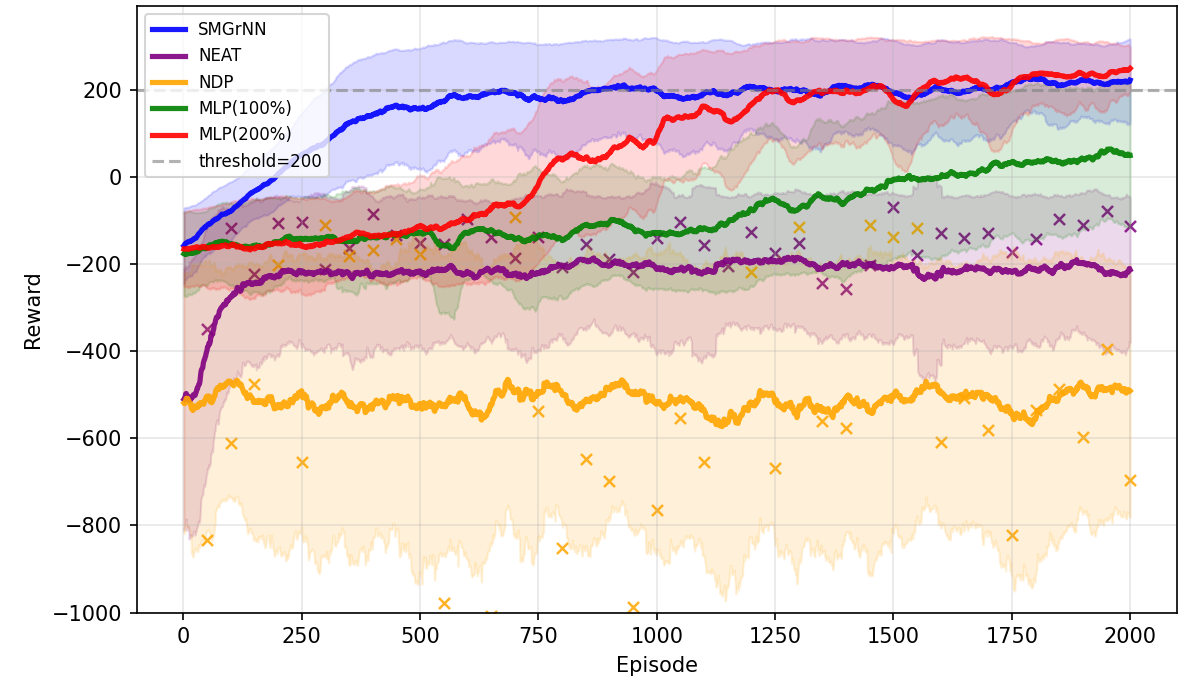}
        \caption{LunarLander-v3 reward}
        \label{fig:baseline_reward_lunarlander}
    \end{subfigure}

    \vspace{0.8em}

    \begin{subfigure}[b]{0.32\textwidth}
        \centering
        \includegraphics[width=\linewidth]{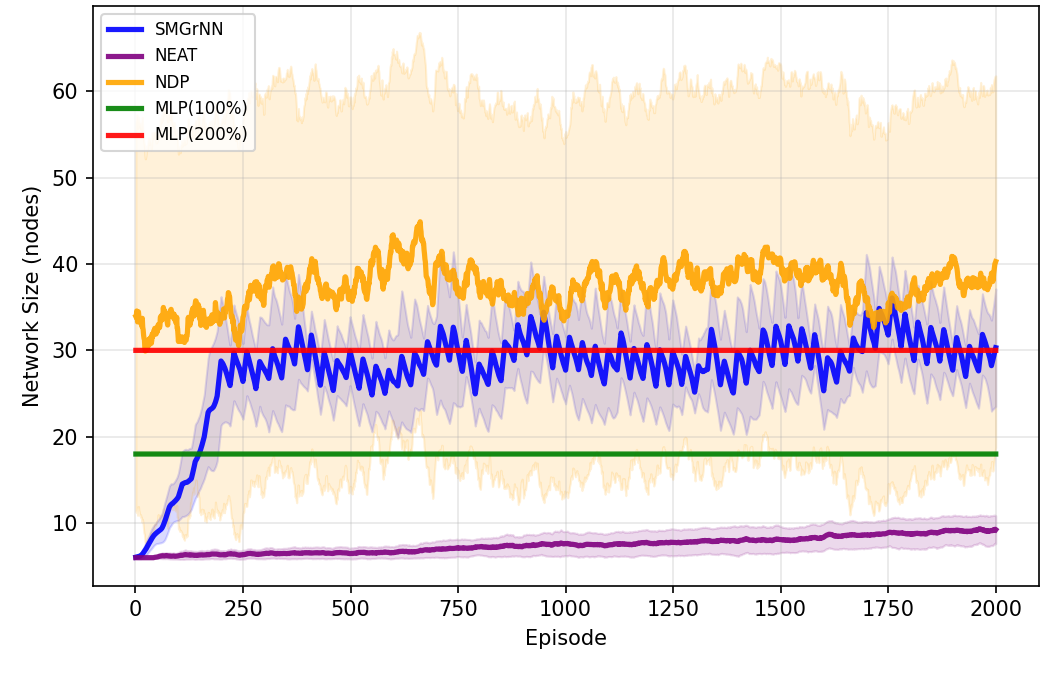}
        \caption{CartPole-v1 size}
        \label{fig:baseline_size_cartpole}
    \end{subfigure}
    \hfill
    \begin{subfigure}[b]{0.32\textwidth}
        \centering
        \includegraphics[width=\linewidth]{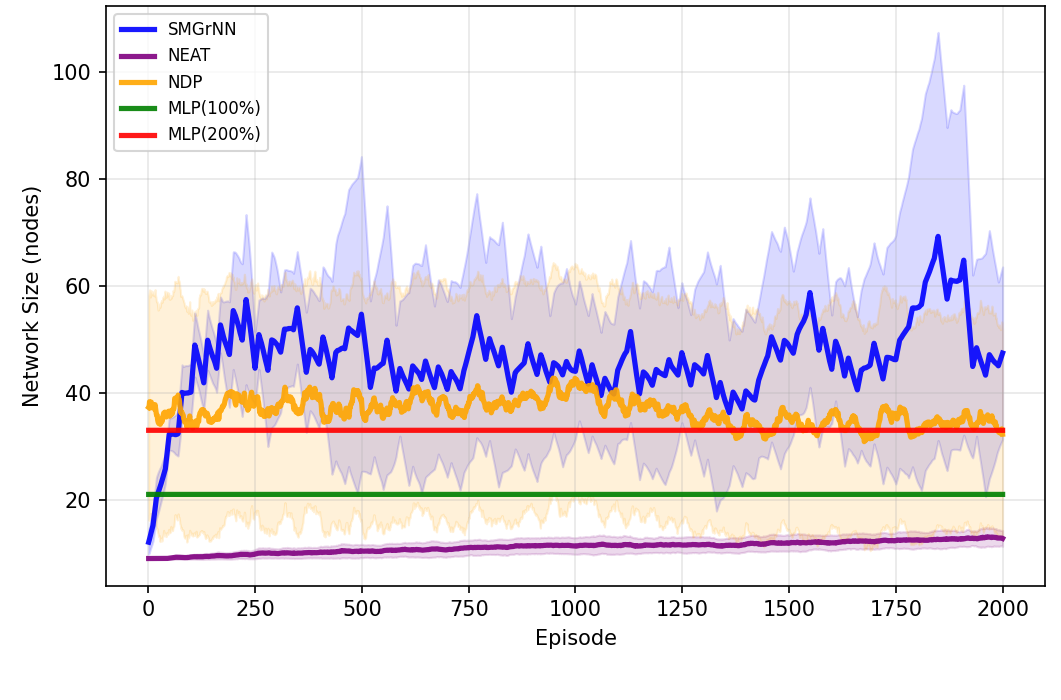}
        \caption{Acrobot-v1 size}
        \label{fig:baseline_size_acrobot}
    \end{subfigure}
    \hfill
    \begin{subfigure}[b]{0.32\textwidth}
        \centering
        \includegraphics[width=\linewidth]{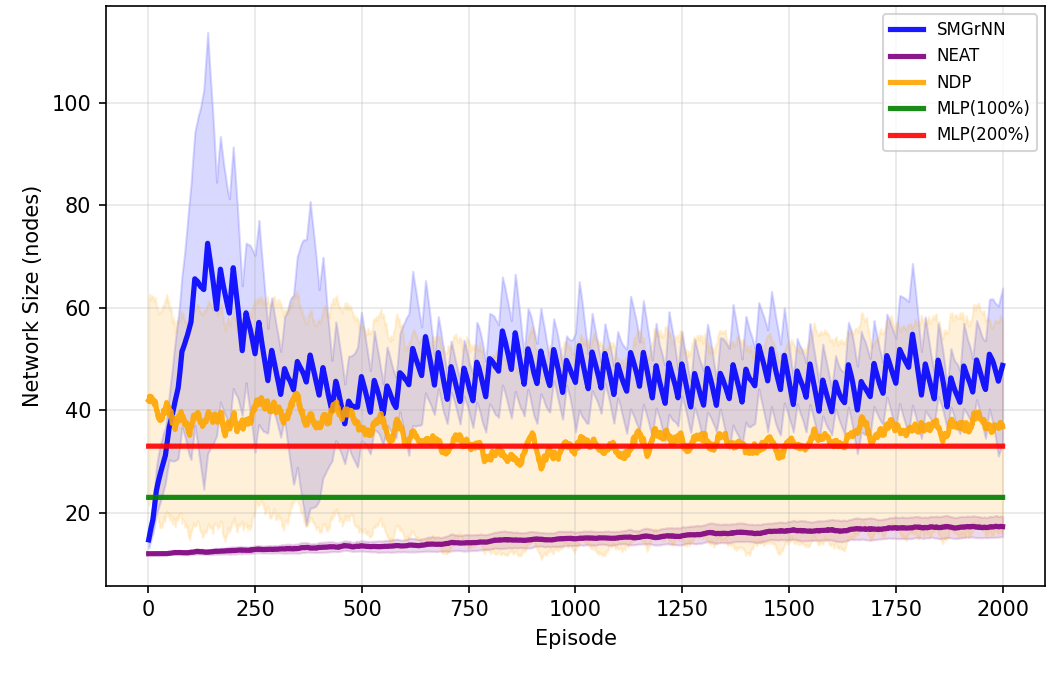}
        \caption{LunarLander-v3 size}
        \label{fig:baseline_size_lunarlander}
    \end{subfigure}

    \caption{Comparison with fixed-topology and dynamic-topology baselines.
    Top row: reward evolution for SMGrNN, MLP baselines, NEAT, and NDP.
    Solid curves show smoothed mean returns over 10 seeds for the
    gradient-trained models and the smoothed evaluation trajectory for the
    evolutionary baselines. Shaded regions indicate across-seed variability,
    and dashed horizontal lines mark task-specific reward thresholds. For
    NEAT and NDP, cross markers indicate the performance of the best
    individual selected in each evolutionary generation. Bottom row:
    network-size evolution and final network sizes. SMGrNN expands to
    task-dependent graph sizes; NEAT exhibits approximately linear
    structural growth; NDP often produces larger and more variable
    structures. MLP baselines have fixed sizes by construction.}
    \label{fig:baseline_comparison}
\end{figure*}

Fig.~\ref{fig:baseline_comparison} summarizes reward evolution and
network-size evolution across fixed-topology and dynamic-topology baselines.
The dynamic-topology baselines should be interpreted with care because NEAT
and NDP are evolutionary or developmental methods rather than
gradient-trained online distillation models. For SMGrNN and the MLP
baselines, the curves represent online training trajectories under the
distillation objective. For NEAT and NDP, the cross markers provide the more
relevant view of search progress, since they indicate the performance of the
best individual selected in each evolutionary generation. The comparison is
therefore intended to assess whether representative topology-search and
developmental-growth baselines can discover competitive controllers under
the same benchmark tasks, rather than to impose identical optimization
dynamics across all methods.

On CartPole-v1, SMGrNN rapidly reaches and maintains near-optimal return.
The fixed MLP baselines also improve, but MLP(100\%) converges more slowly
and MLP(200\%) requires a larger fixed parameter budget to approach similar
performance. NEAT is able to discover near-optimal individuals in several
generations, as shown by generation-best markers close to the task maximum.
However, these high-performing individuals arise through evolutionary
selection rather than through a continuous online policy-improvement
trajectory. NDP shows occasional improvement but does not reliably produce
generation-best individuals in the same high-return regime. Thus, on the
simplest task, dynamic topology search can discover strong candidates, while
SMGrNN provides a continuously trained policy that reaches and maintains
high return with adaptive structural growth.

On Acrobot-v1, SMGrNN again reaches a stable high-return regime under
online training. The fixed MLP baselines improve rapidly at the beginning
but settle at lower return levels than SMGrNN. NEAT also discovers
competitive individuals in some generations, with several generation-best
markers approaching the return range achieved by SMGrNN. This indicates
that evolutionary topology search can identify useful candidate structures on
this task. However, these candidates are selected at the generation level and
do not correspond to a sustained online training trajectory. NDP produces
larger and more variable developmental structures, but its generation-best
performance is less consistently competitive. These results suggest that
dynamic topology can be useful, but the way structural changes are coupled
to learning strongly affects whether discovered structures lead to stable
control performance.

LunarLander-v3 presents a more challenging setting. SMGrNN enters the
positive-reward regime early and reaches high returns substantially earlier
than the fixed MLP baselines. The widest fixed baseline, MLP(200\%), attains
slightly higher late-stage performance, showing that increased fixed capacity
remains a strong baseline on the most difficult task. However, this model
uses a larger fixed parameter budget and does not provide adaptive capacity
control. In contrast to the simpler tasks, the generation-best markers of
NEAT and NDP do not consistently approach the high-return regime reached by
SMGrNN and MLP(200\%). This suggests that generic topology evolution or
developmental growth is less effective on the harder control problem under
the present adapted protocol.

The network-size results further clarify the difference between the methods.
SMGrNN shows task-dependent expansion followed by stabilization: CartPole-v1
induces a relatively small final graph, whereas Acrobot-v1 and
LunarLander-v3 lead to larger structures. This indicates that the SPM does
not simply accumulate parameters, but regulates capacity through growth and
pruning as training proceeds. NEAT exhibits approximately linear structural
growth with relatively small final network sizes. This growth is sufficient
to discover strong individuals on some simpler tasks, but does not scale to
consistently high-performing individuals on LunarLander-v3. NDP often
generates larger and more variable structures, but these structures do not
reliably translate into competitive selected individuals. Static MLPs have
fixed sizes by construction and therefore cannot adjust their capacity during
learning.

Taken together, these results show that dynamic topology alone is not
sufficient to explain the behavior of SMGrNN. Evolutionary and developmental
baselines can occasionally discover high-performing individuals, especially
on simpler tasks, but SMGrNN couples local structural plasticity directly
with online gradient-based policy optimization. This coupling enables
sustained high-return behavior together with task-dependent capacity
regulation, which is the main empirical advantage of the proposed SPM in the
present benchmark protocol.

\subsection{Ablation on structural plasticity}

This section studies the contribution of structural plasticity by
ablating growth and pruning components of the SPM. All models in this ablation start with zero hidden nodes and identical
initial graphs.

\subsubsection{Effect of growth}
\begin{figure*}[t]
    \centering
    \begin{subfigure}[b]{0.32\textwidth}
        \centering
        \includegraphics[width=\linewidth]{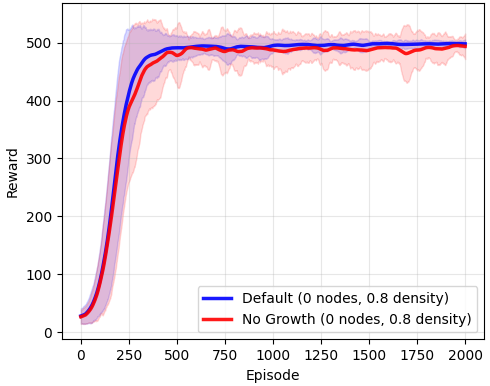}
        \caption{CartPole-v1 reward}
        \label{fig:ablation_growth_cartpole_reward}
    \end{subfigure}
    \hfill
    \begin{subfigure}[b]{0.32\textwidth}
        \centering
        \includegraphics[width=\linewidth]{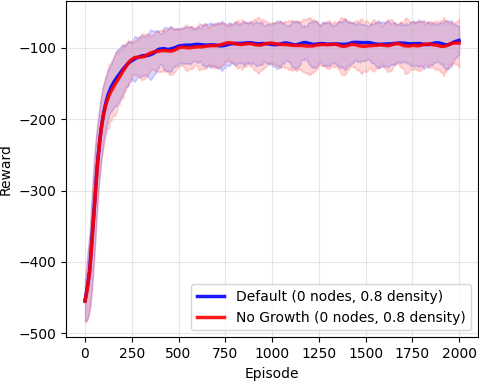}
        \caption{Acrobot-v1 reward}
        \label{fig:ablation_growth_acrobot_reward}
    \end{subfigure}
    \hfill
    \begin{subfigure}[b]{0.32\textwidth}
        \centering
        \includegraphics[width=\linewidth]{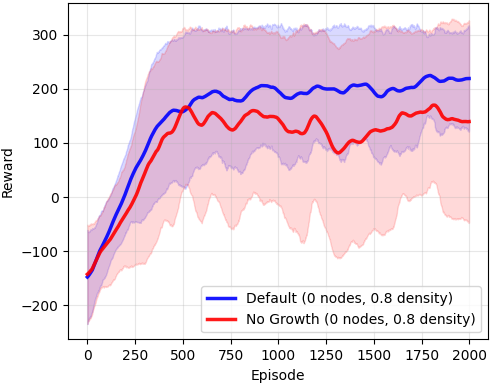}
        \caption{LunarLander-v3 reward}
        \label{fig:ablation_growth_lunar_reward}
    \end{subfigure}

    \vspace{0.6em}

    \begin{subfigure}[b]{0.32\textwidth}
        \centering
        \includegraphics[width=\linewidth]{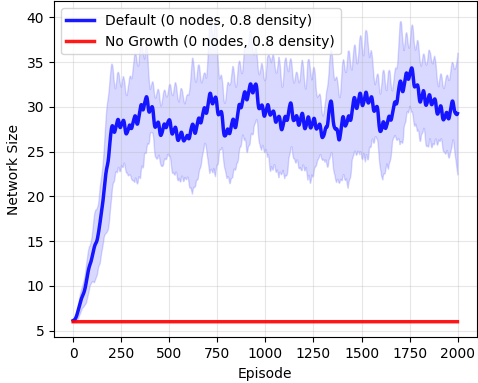}
        \caption{CartPole-v1 hidden nodes}
        \label{fig:ablation_growth_cartpole_size}
    \end{subfigure}
    \hfill
    \begin{subfigure}[b]{0.32\textwidth}
        \centering
        \includegraphics[width=\linewidth]{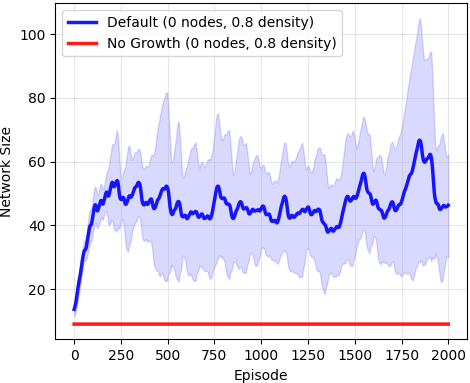}
        \caption{Acrobot-v1 hidden nodes}
        \label{fig:ablation_growth_acrobot_size}
    \end{subfigure}
    \hfill
    \begin{subfigure}[b]{0.32\textwidth}
        \centering
        \includegraphics[width=\linewidth]{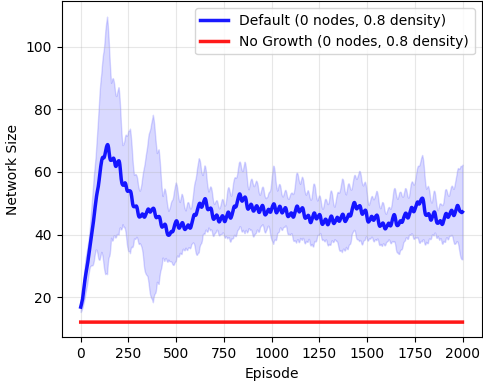}
        \caption{LunarLander-v3 hidden nodes}
        \label{fig:ablation_growth_lunar_size}
    \end{subfigure}

    \caption{Ablation on structural growth. Each panel compares SMGrNN
    (growth enabled) with a static graph in which the Structural Plasticity
    Module is disabled.
    Top row: episodic return over training; bottom row: number of hidden nodes.
    Solid curves show smoothed means, and shaded regions span all 10 runs.}
    \label{fig:ablation_growth}
\end{figure*}

The experimental configuration is summarized in
Table~\ref{tab:ablation_growth}; quantitative results are reported in
Appendix Table~\ref{tab:ablation_growth_stats}, while reward and network
size dynamics are visualized in Fig.~\ref{fig:ablation_growth}.
In the \emph{growth-enabled} condition, the Structural Plasticity Module
(SPM) performs edge-driven and random exploratory growth together with
periodic pruning, as described in Section~\ref{sec:spm}.
In the \emph{static} condition, the SPM is disabled and the graph
remains fixed; only synaptic weights are updated by gradient descent.

\begin{table}[tb]
\centering
\caption{Ablation setup (10 runs per condition).}
\label{tab:ablation_growth}
\footnotesize
\setlength{\tabcolsep}{3pt}
\begin{tabular}{lcccc}
\hline
\textbf{Env.} & \textbf{Episodes} & \textbf{Init hidden} & \textbf{Density} & \textbf{Growth} \\
\hline
CartPole    & 2000 & 0 & 0.8 & Enabled / Disabled \\
Acrobot     & 2000 & 0 & 0.8 & Enabled / Disabled \\
LunarLander & 2000 & 0 & 0.8 & Enabled / Disabled \\
\hline
\end{tabular}
\end{table}

Across CartPole-v1, Acrobot-v1, and LunarLander-v3, enabling structural
growth consistently improves reward stability and reduces run-to-run
variance, most prominently on LunarLander-v3.
Harder tasks also induce larger final architectures under SPM control
(e.g., tens of hidden nodes versus only a small handful in the static
case), indicating that the network expands capacity in response to task
difficulty.
Convergence speed is similar on average between the two conditions but
more consistent when growth is enabled, as reflected in the narrower
spread of convergence episodes in Appendix
Table~\ref{tab:ablation_growth_stats}.
These results suggest that structural growth is an
effective mechanism for matching network capacity to task demands
without manual architectural tuning.

\subsubsection{Effect of pruning}

To assess the role of pruning more reliably, an additional
\emph{growth-only} variant is considered in which the SPM performs
edge-driven and random growth but pruning of weak edges and orphan nodes
is disabled. Both the standard SMGrNN and the growth-only variant are
evaluated over 10 independent runs. This allows us to distinguish the effect
of pruning on task performance from its effect on structural compactness.

Table~\ref{tab:growth_only} summarizes the results and compares the
growth-only variant with the standard SMGrNN configuration.
\begin{table*}[t]
\centering
\caption{Growth-only variant vs.\ standard SMGrNN over 10 runs.
Late reward is the average return over the final quarter of training;
Best reward is the best episodic return observed during training;
Conv.\ episode is the first episode at which the task-specific reward
threshold is reached; Net growth is the increase in hidden-node count
relative to initialization; Parameters is the final number of trainable
parameters.}
\label{tab:growth_only}
\footnotesize
\setlength{\tabcolsep}{4pt}
\begin{tabular}{llccccc}
\hline
\textbf{Env.} & \textbf{Model} & \textbf{Late reward} & \textbf{Best} & \textbf{Conv.\ Ep. }& \textbf{Growth }& \textbf{Params} \\
\hline
CartPole
& Standard
& $497.7 \pm 1.8$
& $500.0 \pm 0.0$
& $147 \pm 15$
& $24.2$
& $84 \pm 16$ \\
& No-prune
& $493.6 \pm 3.6$
& $500.0 \pm 0.0$
& $151 \pm 11$
& $1891.4$
& $6128 \pm 1108$ \\
\hline
Acrobot
& Standard
& $-94.0 \pm 1.9$
& $-62.3 \pm 0.5$
& $80 \pm 12$
& $35.4$
& $119 \pm 39$ \\
& No-prune
& $-90.6 \pm 1.6$
& $-62.4 \pm 0.8$
& $83 \pm 13$
& $2408.7$
& $6761 \pm 998$ \\
\hline
LunarLander
& Standard
& $207.6 \pm 33.3$
& $321.1 \pm 5.3$
& $231 \pm 16$
& $34.0$
& $140 \pm 40$ \\
& No-prune
& $211.3 \pm 36.7$
& $315.7 \pm 6.0$
& $245 \pm 22$
& $5535.3$
& $16707 \pm 4846$ \\
\hline
\end{tabular}
\end{table*}

\begin{figure}[ht]
    \centering
    \begin{subfigure}[b]{0.48\linewidth}
        \centering
        \includegraphics[width=\linewidth]{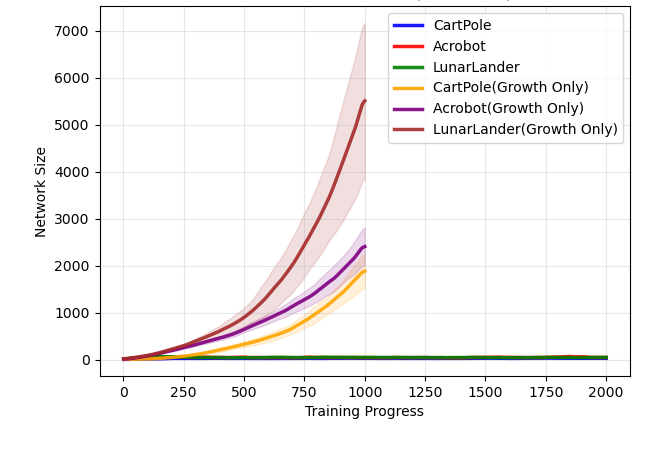}
        \caption{Node growth across tasks}
        \label{fig:growthonly_growth_all}
    \end{subfigure}
    \hfill
    \begin{subfigure}[b]{0.48\linewidth}
        \centering
        \includegraphics[width=\linewidth]{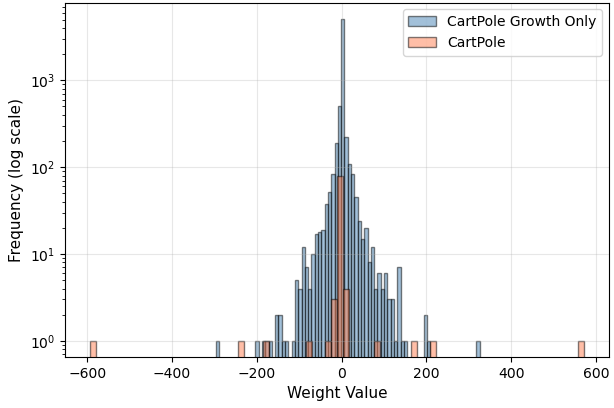}
        \caption{CartPole-v1 log-scale histogram}
        \label{fig:growthonly_cartpole_log}
    \end{subfigure}

    \vspace{0.6em}

    \begin{subfigure}[b]{0.48\linewidth}
        \centering
        \includegraphics[width=\linewidth]{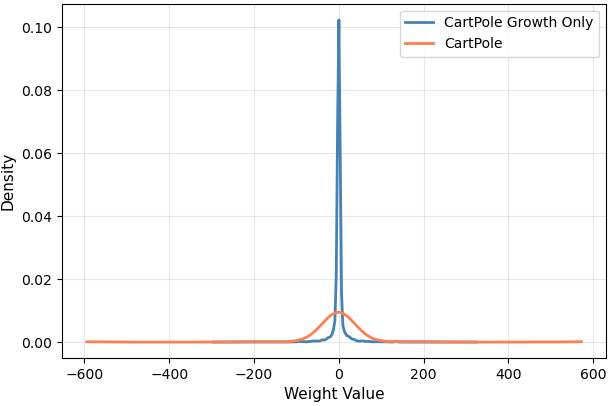}
        \caption{CartPole-v1 weight PDF}
        \label{fig:growthonly_cartpole_pdf}
    \end{subfigure}
    \hfill
    \begin{subfigure}[b]{0.48\linewidth}
        \centering
        \includegraphics[width=\linewidth]{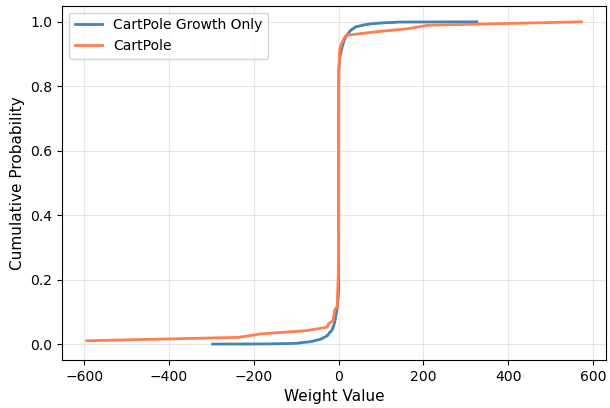}
        \caption{CartPole-v1 weight CDF}
        \label{fig:growthonly_cartpole_cdf}
    \end{subfigure}

    \caption{Growth-only variant vs.\ SMGrNN.
    Panel~(a) shows the number of nodes over training for all
    environments.
    Panels~(b)--(d) show final weight distributions on CartPole-v1:
    log-scale histogram, probability density function (PDF), and
    cumulative distribution function (CDF).}
    \label{fig:growthonly_main}
\end{figure}

Across all three tasks, the growth-only variant achieves late-training
rewards, best rewards, and convergence episodes that are broadly comparable
to those of standard SMGrNN. On CartPole-v1, both variants reach the
maximum episodic return during training, and their late-training rewards
differ only slightly. On Acrobot-v1, the growth-only variant achieves a
slightly higher late-training reward while reaching essentially the same best
reward. On LunarLander-v3, the two variants also remain in a comparable
performance range, although the growth-only model reaches the task threshold
slightly later and attains a slightly lower best reward.

However, this comparable task performance comes at the cost of explosive
structural expansion. On CartPole-v1, disabling pruning increases the final
parameter count from $84 \pm 16$ to $6128 \pm 1108$, while net growth rises
from $24.2$ to $1891.4$. On Acrobot-v1, the final parameter count increases
from $119 \pm 39$ to $6761 \pm 998$, with net growth increasing from
$35.4$ to $2408.7$. The effect is even stronger on LunarLander-v3, where
the growth-only variant reaches $16707 \pm 4846$ parameters and a net
growth of $5535.3$, compared with only $140 \pm 40$ parameters and a net
growth of $34.0$ for standard SMGrNN.

Fig.~\ref{fig:growthonly_main} shows how this excess capacity arises:
panel~\ref{fig:growthonly_growth_all} reveals unchecked growth in the
number of hidden nodes, and the CartPole-v1 weight distributions in
panels~\ref{fig:growthonly_cartpole_log}--\ref{fig:growthonly_cartpole_cdf}
indicate a much broader spectrum of weights with many small-magnitude
connections under growth-only training. In contrast, standard SMGrNN prunes
low-impact edges and removes orphan nodes, concentrating capacity on a
more compact active subgraph. These results indicate that pruning is not
primarily required to make learning possible; rather, it is critical for
preventing uncontrolled network expansion and for keeping structural
plasticity compact and computationally tractable while preserving comparable
control performance.

\subsubsection{Sensitivity to initial topology}

Sensitivity to the initial graph is examined by varying the initial
hidden-node count in $\{0,5,10\}$ while keeping the SPM and all other
hyperparameters fixed.
Table~\ref{tab:topology_setup} summarizes the configuration, aggregate
metrics are reported in Appendix Table~\ref{tab:topology_summary}, and
Fig.~\ref{fig:topology_results} shows the evolution of network size
and average growth over training.

\begin{table}[tb]
\centering
\caption{Setup for the initial-topology sensitivity experiment (10 runs per configuration).}
\label{tab:topology_setup}
\footnotesize
\setlength{\tabcolsep}{3pt}
\begin{tabular}{lcccc}
\hline
\textbf{Environment} & \textbf{Episodes} & \textbf{Initial hidden} & \textbf{Edge density} & \textbf{Growth} \\
\hline
CartPole    & 2000 & 0 / 5 / 10 & 0.8 & Enabled \\
Acrobot     & 2000 & 0 / 5 / 10 & 0.8 & Enabled \\
LunarLander & 2000 & 0 / 5 / 10 & 0.8 & Enabled \\
\hline
\end{tabular}
\end{table}

\begin{figure*}[t]
    \centering
    \begin{subfigure}[b]{0.32\textwidth}
        \centering
        \includegraphics[width=\linewidth]{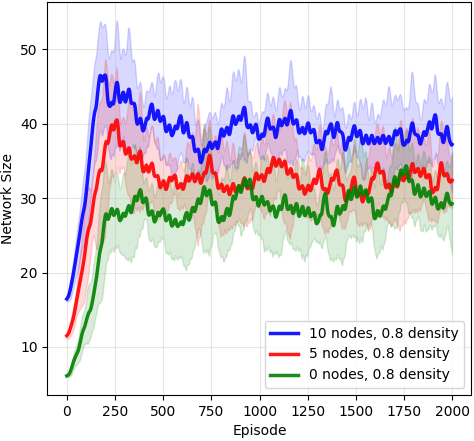}
        \caption{CartPole-v1 size}
        \label{fig:topology_cartpole_size}
    \end{subfigure}
    \hfill
    \begin{subfigure}[b]{0.32\textwidth}
        \centering
        \includegraphics[width=\linewidth]{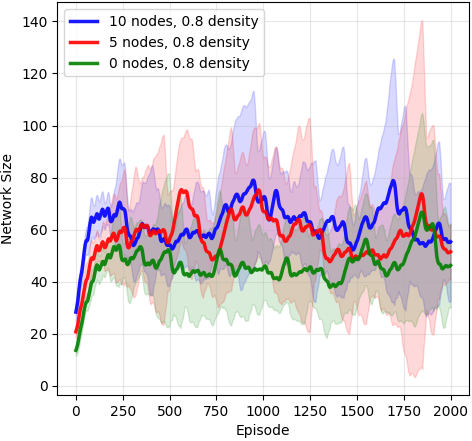}
        \caption{Acrobot-v1 size}
        \label{fig:topology_acrobot_size}
    \end{subfigure}
    \hfill
    \begin{subfigure}[b]{0.32\textwidth}
        \centering
        \includegraphics[width=\linewidth]{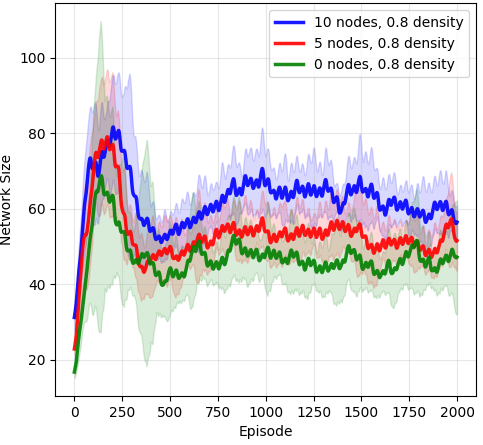}
        \caption{LunarLander-v3 size}
        \label{fig:topology_lunar_size}
    \end{subfigure}

    \vspace{0.6em}

    \begin{subfigure}[b]{0.32\textwidth}
        \centering
        \includegraphics[width=\linewidth]{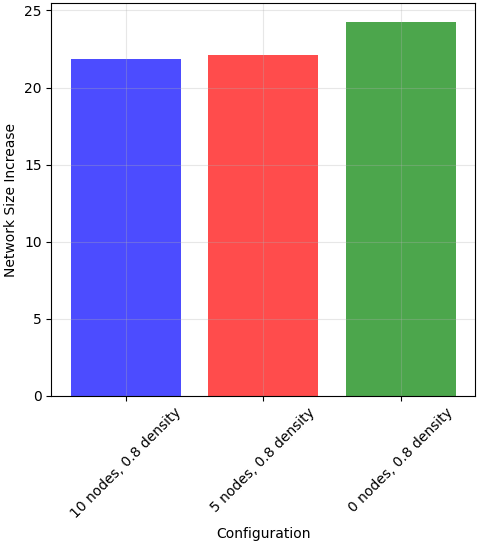}
        \caption{CartPole-v1 average growth}
        \label{fig:topology_cartpole_growth}
    \end{subfigure}
    \hfill
    \begin{subfigure}[b]{0.32\textwidth}
        \centering
        \includegraphics[width=\linewidth]{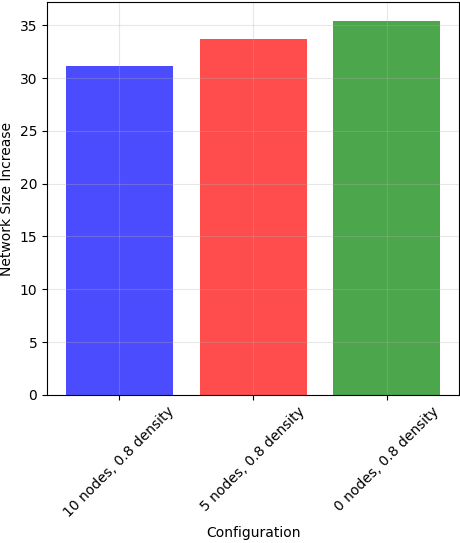}
        \caption{Acrobot-v1 average growth}
        \label{fig:topology_acrobot_growth}
    \end{subfigure}
    \hfill
    \begin{subfigure}[b]{0.32\textwidth}
        \centering
        \includegraphics[width=\linewidth]{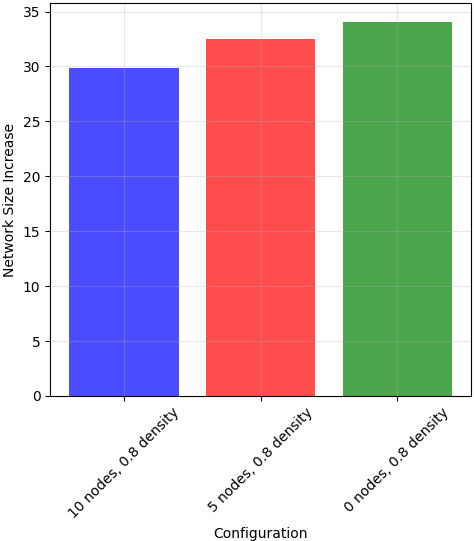}
        \caption{LunarLander-v3 average growth}
        \label{fig:topology_lunar_growth}
    \end{subfigure}

    \caption{Sensitivity to initial topology.
    Each panel averages over 10 runs with initial hidden-node counts
    $0$, $5$, and $10$ under identical SPM hyperparameters.
    Top row: number of hidden nodes over training; bottom row: average
    growth per episode.}
    \label{fig:topology_results}
\end{figure*}

Larger initial capacity induces \emph{less} additional growth, and
sparser initial graphs trigger stronger early expansion.
Across initializations, final hidden-node counts partially converge to a
similar range in each environment, indicating compensatory scaling by
the SPM.
Late-training rewards are nearly identical on CartPole-v1 and
Acrobot-v1, and only modest gains are observed for richer initial
topologies on the more challenging LunarLander-v3
(see Appendix Table~\ref{tab:topology_summary}).
Overall, SMGrNN shows low sensitivity to the initial hidden size, as
over- and under-parameterized starting graphs are corrected online by
growth and pruning.

\subsubsection{Task difficulty and structural scaling}

\begin{figure}[tb]
    \centering
    \begin{subfigure}[b]{0.45\textwidth}
        \centering
        \includegraphics[width=\linewidth]{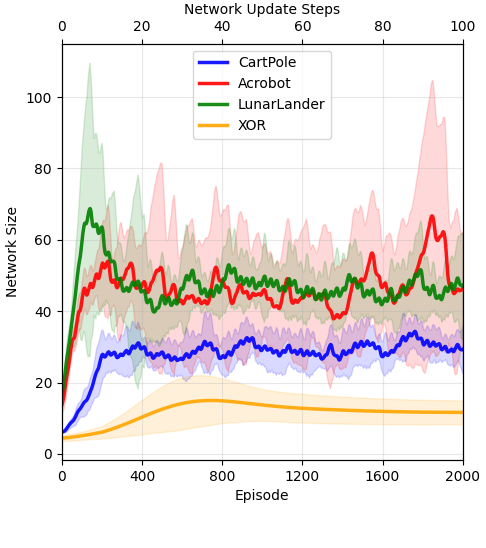}
        \caption{Network size over training}
        \label{fig:difficulty_size_traj}
    \end{subfigure}
    \hfill
    \begin{subfigure}[b]{0.45\textwidth}
        \centering
        \includegraphics[width=\linewidth]{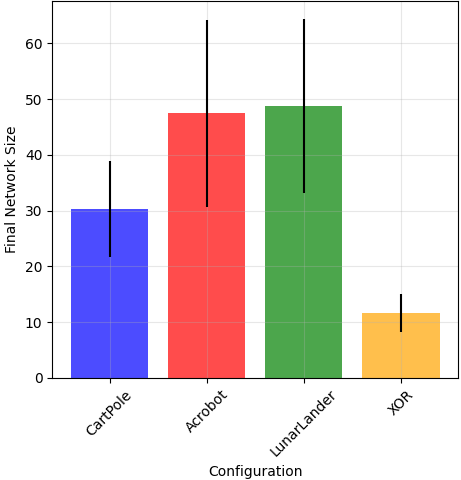}
        \caption{Final network size}
        \label{fig:difficulty_size_final}
    \end{subfigure}
    \caption{Scaling of SMGrNN capacity with task difficulty.
    Left: hidden-node trajectories; right: final hidden-node counts.}
    \label{fig:difficulty_results}
\end{figure}

Structural scaling with task difficulty is investigated by training
SMGrNN on four tasks under identical SPM hyperparameters:
a trivial supervised XOR classification problem, CartPole-v1 (easy),
Acrobot-v1 (moderate), and LunarLander-v3 (harder).
Fig.~\ref{fig:difficulty_results} shows the evolution of hidden-node
counts and the corresponding final sizes, and Appendix
Table~\ref{tab:difficulty_summary} reports numerical summaries.

Structural scale follows task complexity.
On the XOR problem the network grows minimally and quickly stabilizes at
a very small size, reflecting the low representational demand.
On CartPole-v1 the topology expands moderately from its minimal seed and
stabilizes at a small network.
On Acrobot-v1 the architecture grows further and settles at an
intermediate scale.
On LunarLander-v3 the network continues to grow for longer and converges
to the largest size among the four tasks
(Appendix Table~\ref{tab:difficulty_summary}).
These patterns indicate that, under a single set of SPM hyperparameters,
SMGrNN allocates capacity in proportion to task difficulty without
manual adjustment of the architecture.

\subsection{Qualitative analysis of final topologies}
\label{sec:final_topology}

\begin{figure*}[t]
    \centering
    \begin{subfigure}[b]{0.32\textwidth}
        \centering
        \includegraphics[width=\linewidth]{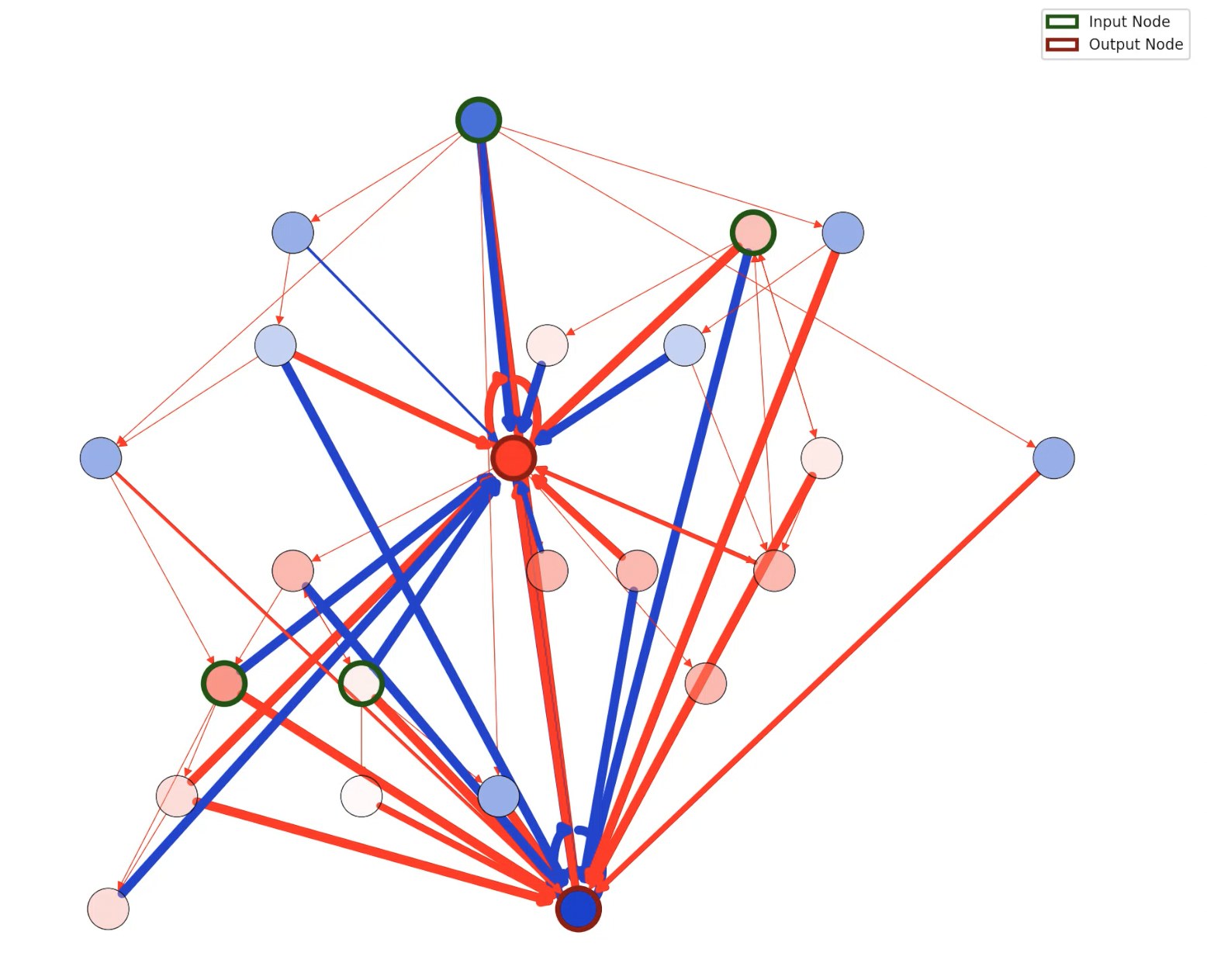}
        \caption{CartPole-v1}
        \label{fig:final_topology_cartpole}
    \end{subfigure}
    \hfill
    \begin{subfigure}[b]{0.32\textwidth}
        \centering
        \includegraphics[width=\linewidth]{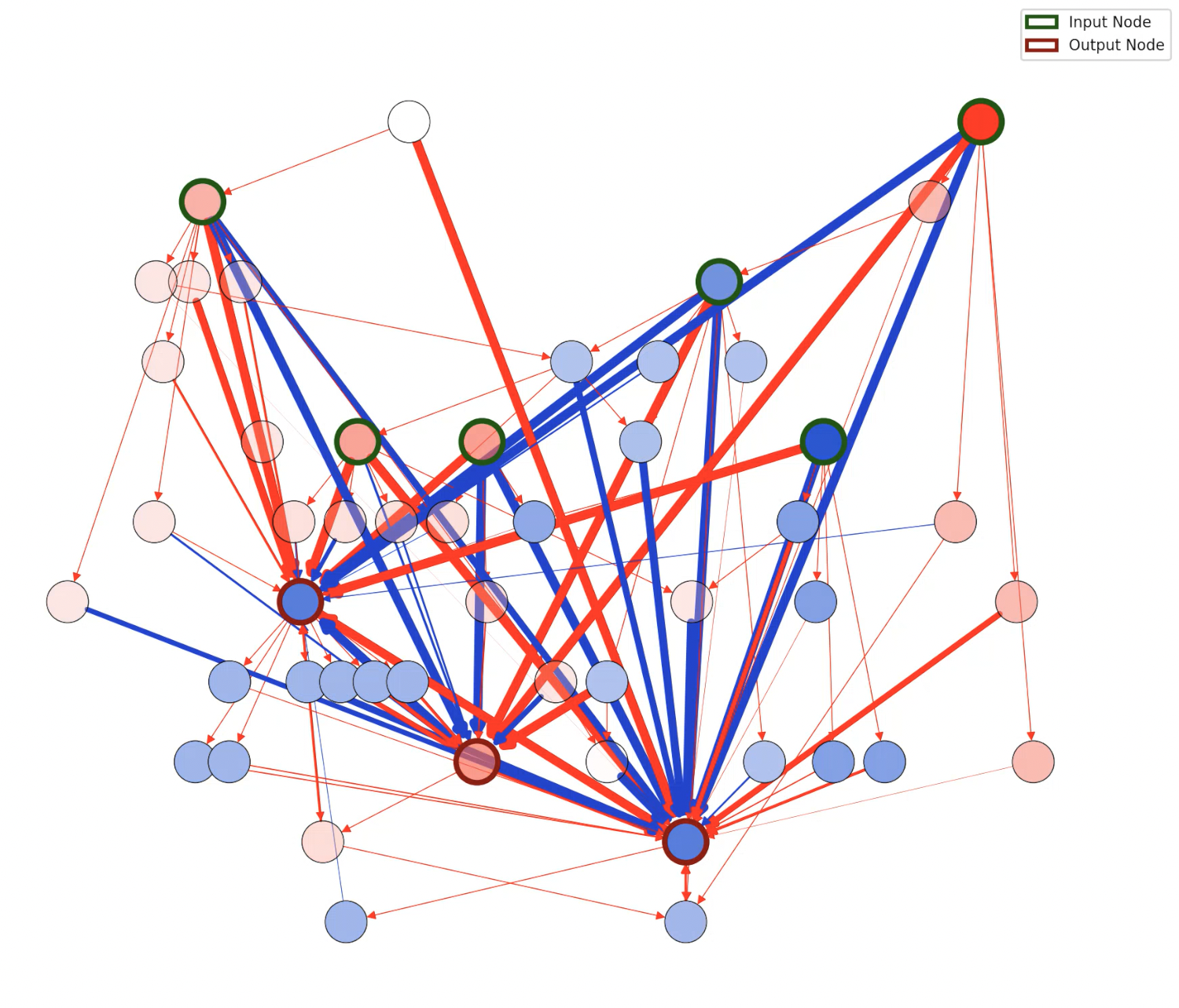}
        \caption{Acrobot-v1}
        \label{fig:final_topology_acrobot}
    \end{subfigure}
    \hfill
    \begin{subfigure}[b]{0.32\textwidth}
        \centering
        \includegraphics[width=\linewidth]{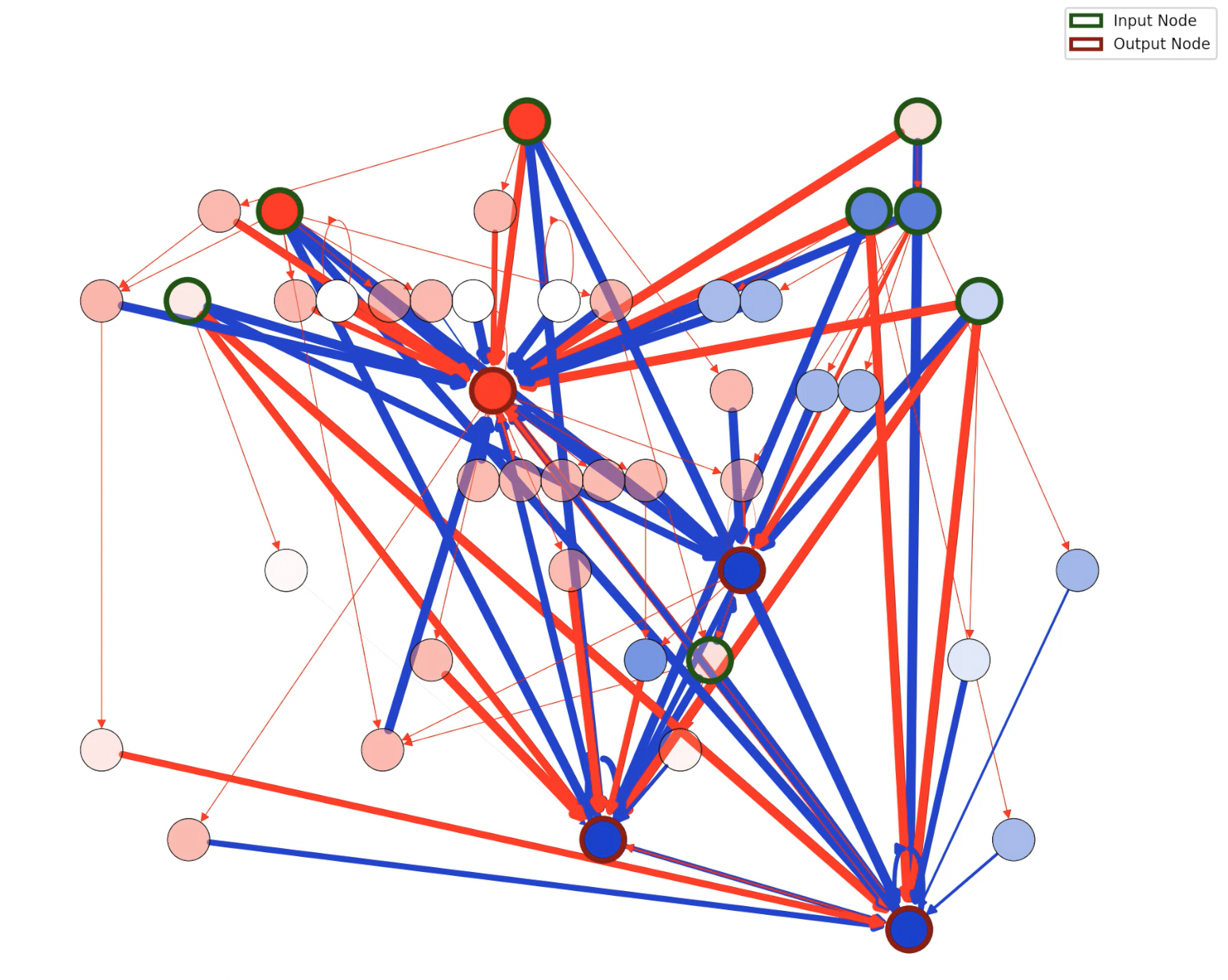}
        \caption{LunarLander-v3}
        \label{fig:final_topology_lunarlander}
    \end{subfigure}
    \caption{Representative final SMGrNN topologies after training on
    CartPole-v1, Acrobot-v1, and LunarLander-v3. Input nodes are outlined
    in green and output nodes in red. Edge color and width indicate signed
    weight magnitude. The learned graphs are sparse, directed, and
    non-layered, illustrating how local structural plasticity produces
    task-dependent topologies that differ substantially from a fixed
    multilayer perceptron.}
    \label{fig:final_topologies}
\end{figure*}

To qualitatively illustrate the structures produced by online structural
plasticity, Fig.~\ref{fig:final_topologies} shows representative final
topologies learned by SMGrNN on CartPole-v1, Acrobot-v1, and
LunarLander-v3. These graphs are no longer layerwise multilayer
perceptrons: relay-node insertion, exploratory edge growth, and pruning
produce sparse directed topologies with skip connections, feedback edges,
and task-dependent connectivity patterns.

Several qualitative trends are visible. First, the learned structures are
sparse rather than densely layered, indicating that the SPM does not simply
accumulate parameters but selectively preserves a compact active subgraph.
Second, the final topologies differ substantially across tasks, which is
consistent with the quantitative network-size scaling reported in
Fig.~\ref{fig:difficulty_results}. Simpler tasks such as CartPole-v1 tend to
yield smaller and more centralized structures, whereas more challenging
tasks such as Acrobot-v1 and LunarLander-v3 produce larger graphs with
richer long-range connectivity. Third, the final networks contain motifs that
cannot be represented as a standard single-hidden-layer MLP, including
relay-mediated pathways, non-layered directed links, and feedback
connections.

These visualizations complement the quantitative analyses in
Sections~4.2--4.3 by showing directly that the resulting architectures are
task-dependent, non-layered, and shaped by the combined action of growth
and pruning.

\section{Discussion}

\subsection{Summary of contributions and empirical findings}

The present study investigated a simple form of local structural
plasticity for control policies, instantiated in the
SMGrNN.
SMGrNN augments a standard, gradient-trained policy network with a
SPM that adjusts the graph topology
during training based on short-term statistics of edge-wise weight updates 
and derivatives.
The SPM uses only locally available information collected in rolling
history buffers to decide when and where to grow or prune hidden nodes
and connections, while synaptic weights are optimized by conventional
gradient descent.

Experiments on three benchmark control tasks showed that this form of
local structural plasticity provides competitive performance relative to both
fixed-topology and adapted dynamic-topology baselines. With a single set of
SPM hyperparameters, SMGrNN matches or exceeds the returns of
parameter-matched static MLPs on CartPole-v1 and Acrobot-v1, while
reaching strong performance earlier than the wider MLP baseline on
LunarLander-v3 (Fig.~\ref{fig:baseline_comparison}). The added NEAT
and NDP comparisons further show that dynamic topology alone does not
fully explain the observed behavior. Evolutionary and developmental
baselines can discover strong individuals in simpler tasks, but SMGrNN
couples structural adaptation with online gradient-based policy optimization,
yielding sustained high-return behavior together with task-dependent
capacity regulation (Fig.~\ref{fig:baseline_comparison}).

Ablation studies further clarified the role of structural plasticity.
Disabling growth and pruning reduces SMGrNN to a fixed graph and yields
more fragile performance that is sensitive to the initial architecture
(Fig.~\ref{fig:ablation_growth}, Table~\ref{tab:ablation_growth_stats}).
Conversely, a growth-only variant that removes pruning achieves comparable
late-training rewards but allows the network to expand to orders of
magnitude more parameters, producing broad weight distributions with many
small-magnitude connections (Fig.~\ref{fig:growthonly_main},
Table~\ref{tab:growth_only}). Additional analyses showed that SMGrNN is
robust to the initial number of hidden nodes, with final capacities partially
converging across initializations (Fig.~\ref{fig:topology_results},
Table~\ref{tab:topology_summary}), and that emergent network size scales
systematically with task difficulty across XOR, CartPole-v1, Acrobot-v1,
and LunarLander-v3 (Fig.~\ref{fig:difficulty_results},
Table~\ref{tab:difficulty_summary}). Taken together, these results indicate
that the SPM enables a single controller to self-adjust its effective capacity
to task demands, with minimal manual tuning of architectural hyperparameters.

\subsection{Role of local structural plasticity}

The proposed SPM implements structural plasticity through local
statistical signals rather than explicit global objectives.
Decisions to grow or prune structure are based on edge-wise update 
statistics within a short temporal window,
without direct access to the global reinforcement signal or the full
loss landscape.
In particular, edge-driven growth is triggered by fluctuating
derivatives, suggesting insufficiently resolved structure along certain
paths, whereas pruning targets weak and near-static edges and removes
orphan nodes.
This separates the \emph{structural} rule, which acts on intrinsic
signals, from the \emph{synaptic} rule, which in this work is provided
by backpropagation.

This separation places SMGrNN in a distinct position relative to most
dynamic architectures in continual and multi-task learning, where
capacity expansion is often orchestrated at the task level or via
heuristics tied to validation performance.
Here, structural adaptation proceeds continuously and at a finer
granularity, with small changes accumulated over time as the policy
trains.
In the present work, this separation should be interpreted conservatively. The experiments only establish that local structural adaptation can be beneficial when combined with standard gradient-based synaptic optimization. They do not show that the same structural rules remain effective under Hebbian, spike-based, or other fully local synaptic learning rules. The main significance of the current result is therefore the independent contribution of local structural plasticity within a conventional training setting, rather than a validated general framework for all local learning regimes.

\subsection{Limitations and future directions}

Several limitations of the current study delimit the scope of the
conclusions.

First, the evaluation is restricted to single-task policy distillation and simple supervised classification (XOR), rather than full continual learning benchmarks involving sequences of distinct tasks or explicit tests of catastrophic forgetting.
The reported structural adaptation therefore reflects within-task
capacity adjustment, not long-term maintenance of multiple skills.
In other words, the focus here is on within-task capacity adaptation, not on catastrophic forgetting across task sequences; explicit continual learning is left to future work.

Second, although structural edits may produce feedback edges and other
non-layered motifs, the current implementation handles such topologies
through a fixed number of synchronous message-passing iterations within
each forward call, with gradients propagated only through this finite
unrolling. The empirical evaluation in this paper covers two limited
regimes: \(K=1\) for the control tasks, where recurrent connectivity is
mediated mainly through cross-call state, and \(K=3\) for the XOR task,
where limited within-call iterative propagation is used because no cross-call
state is available. The present study therefore does not provide a general
treatment of arbitrary graph topologies under unrestricted recurrent credit
assignment; rather, it studies local structural plasticity in a graph-state
network under these restricted computational settings.

Third, the SPM itself is hand-designed. Its thresholds, sampling ratios,
and update schedules are selected by preliminary tuning on a small set of
environments rather than learned automatically. Although the final
configuration is shared across tasks in the present experiments, the current
study does not provide an adaptive mechanism for selecting these structural
hyperparameters. Learning or meta-optimizing the growth and pruning
policy remains an important direction for future work.

Fourth, the current implementation is not optimized for computational
efficiency. Graph-state message passing, history-buffer maintenance, and
structural editing are implemented in a prototype form rather than as fully
vectorized or batched operations. Preliminary timing checks indicate that
the main overhead arises from graph execution and SPM structural updates
rather than from the backward pass itself. More efficient vectorized graph
execution, batched structural updates, and optimized history-buffer
management remain future engineering directions.

A further limitation is that the present study does not isolate the
contribution of criterion-driven relay insertion from that of random
exploratory growth, nor does it compare relay insertion against alternative
local structural edit operators. A systematic comparison among different
edit operators remains for future work.

These limitations point to several natural directions for future work.
A first step would be to replace or augment gradient-based synaptic
updates with local learning rules, such as Hebbian or forward-forward style objectives in rate-based networks, or spike-timing-dependent plasticity in spiking implementations.
Because the SPM operates purely on local edge-wise update statistics, the same structural mechanism could, in principle, be
combined with such rules to obtain controllers in which both structure and weights are governed by local plasticity modules.
A second direction is to move from single-task distillation to settings with explicit task sequences or non-stationary environments, in order to test whether SPM-driven growth and pruning can allocate and preserve substructures that support multiple skills while mitigating interference.
A third avenue is to make the SPM itself adaptive, for example by
meta-learning or reinforcement learning over growth and pruning
hyperparameters, so that structural policies are tuned by performance signals rather than fixed by hand.
Exploring these directions may help to turn local structural plasticity from a useful architectural heuristic into a more general principle for adaptive, resource-aware control.

\bibliographystyle{IEEEtran}
\bibliography{SMGrNN}

\appendix
\renewcommand{\thetable}{A.\Roman{table}}
\setcounter{figure}{0}
\setcounter{table}{0}
\counterwithin{figure}{section}

\section{Supplementary Materials}
\label{app:supplementary}

This appendix provides supplementary materials supporting the analyses in the
main text, including hyperparameter settings, hyperparameter selection
protocols, sensitivity analyses, and additional quantitative results. Unless
otherwise noted, all statistics are averaged over multiple runs with different
random seeds.
\subsection{Optimizer and learning-rate sensitivity}
\label{app:optimizer_lr_sensitivity}

To examine whether the normalized instability criterion is purely an
optimizer-scale artifact, we performed a small sensitivity study on
CartPole-v1 while keeping all SPM hyperparameters fixed to the paper
defaults. Each setting was evaluated with 5 random seeds and 1000 episodes
per seed. We first compared several optimizers at a fixed learning rate of
0.01, and then varied the learning rate of Adam over
\(\{0.001, 0.003, 0.01, 0.03\}\).

Across all tested optimizers, growth and pruning remained active, indicating
that replacing the optimizer does not disable the SPM trigger. However,
optimizer choice does affect the magnitude and stability of structural
adaptation. In particular, RMSprop produces excessive structural expansion
and very high turnover relative to Adam. Under Adam, varying the learning
rate from 0.001 to 0.03 also preserves active structural adaptation, but
larger learning rates increase final network size and churn. Overall, these
results suggest that the normalized instability criterion is functional under
optimizer and learning-rate changes, while Adam with a moderate learning
rate provides the most stable performance--structure trade-off in this study.

\begin{table*}[t]
\centering
\caption{CartPole-v1 optimizer sensitivity of the SPM (5 seeds, 1000 episodes per setting). All runs use the paper-default SPM hyperparameters and a fixed learning rate of 0.01.}
\label{tab:optimizer_sensitivity_cartpole}
\small
\begin{tabular}{lccc}
\hline
\textbf{Optimizer} & \textbf{Late return} & \textbf{Final nodes} & \textbf{Churn ratio} \\
\hline
Adam & $495.61 \pm 1.44$ & $29.80 \pm 5.71$ & $44.08 \pm 9.06$ \\
AdamW & $477.93 \pm 3.19$ & $32.20 \pm 3.60$ & $50.15 \pm 5.57$ \\
RMSprop & $484.76 \pm 3.42$ & $814.80 \pm 498.08$ & $68.57 \pm 36.56$ \\
SGD ($m=0.9$) & $471.99 \pm 9.84$ & $50.40 \pm 7.91$ & $58.20 \pm 5.49$ \\
\hline
\end{tabular}
\end{table*}

\begin{table*}[t]
\centering
\caption{CartPole-v1 learning-rate sensitivity of the SPM under Adam (5 seeds, 1000 episodes per setting).}
\label{tab:lr_sensitivity_cartpole}
\small
\begin{tabular}{lccc}
\hline
\textbf{Learning rate} & \textbf{Late return} & \textbf{Final nodes} & \textbf{Churn ratio} \\
\hline
0.001 & $496.61 \pm 0.92$ & $12.80 \pm 0.40$ & $36.09 \pm 4.00$ \\
0.003 & $498.46 \pm 0.23$ & $18.60 \pm 4.76$ & $36.72 \pm 11.50$ \\
0.01  & $494.60 \pm 2.74$ & $31.20 \pm 3.97$ & $41.60 \pm 6.65$ \\
0.03  & $489.36 \pm 6.73$ & $49.60 \pm 4.45$ & $52.07 \pm 6.62$ \\
\hline
\end{tabular}
\end{table*}

Here, the churn ratio is defined as the total number of structural add/remove
operations divided by the magnitude of the final net structural change. A
larger churn ratio therefore indicates greater structural turnover relative to
the final retained topology.

\subsection{Hyperparameter selection and sensitivity}
\label{app:hparam_selection}

This subsection summarizes the Structural Plasticity Module (SPM)
hyperparameters used in the main experiments, together with the protocol used
to select them. The same final configuration is applied across tasks in order
to highlight how structural adaptation emerges from local statistics rather
than from task-specific tuning. The numerical values of the final shared
defaults are listed in Table~\ref{tab:spm_hyperparams}.

\begin{table*}[t]
\centering
\caption{Hyperparameters of the SPM used in all experiments.}
\label{tab:spm_hyperparams}
\begin{tabular}{lll}
\hline
Symbol & Name & Value \\
\hline
$T$ & Temporal window length (steps) & $100$ \\
$p_{\mathrm{rand}}$ & Random growth probability & $0.25$ \\
$\rho_{\mathrm{rand}}$ & Random growth ratio & $0.01$ \\
$\eta_{\mathrm{prune}}$ & Pruned weak-edge fraction & $0.8$ \\
$s$ & Pruning period (steps) & $30$ \\
$\tau_w$ & Weight threshold for weak edges & $0.1$ \\
$\tau_{\Delta}$ & Derivative threshold for static edges & $10^{-4}$ \\
$\lambda$ & Zero-crossing interval width & $0.5$ \\
$\theta_{\mathrm{var}}$ & Threshold for normalized fluctuation magnitude & $0.1$ \\
\hline
\end{tabular}
\end{table*}

Only the four primary SPM hyperparameters
$(T,\lambda,\theta_{\mathrm{var}},p_{\mathrm{rand}})$ were included in the
actual tuning search. These parameters were selected because they directly
determine the temporal sensitivity of the update statistics, the instability
criterion for edge-driven growth, and the frequency of exploratory growth,
and therefore define the core structural adaptation behaviour of the SPM.
The remaining SPM hyperparameters
$(\rho_{\mathrm{rand}},\eta_{\mathrm{prune}},s,\tau_w,\tau_\Delta)$ were held
fixed at conservative defaults because they mainly regulate secondary aspects
of structural editing, such as pruning frequency, pruning aggressiveness, and
random-growth batch size.

Phase~1 used a full $3^4$ grid search on CartPole-v1 over
$T \in \{50,100,200\}$,
$\lambda \in \{0.25,0.5,1.0\}$,
$\theta_{\mathrm{var}} \in \{0.05,0.1,0.2\}$, and
$p_{\mathrm{rand}} \in \{0.1,0.25,0.5\}$.
Each configuration was evaluated with 5 random seeds over 500 episodes and
ranked by late-training reward, defined as the mean episodic return over the
final quarter of training, with ties broken first by lower across-seed
variance and then by smaller final network size. Under this protocol, the
best CartPole-v1 configuration was
$T=50$, $\lambda=0.5$, $\theta_{\mathrm{var}}=0.05$, and
$p_{\mathrm{rand}}=0.25$, achieving a late-training reward of
$494.26 \pm 3.57$.

Phase~2 performed one-factor-at-a-time sensitivity analyses on CartPole-v1
and Acrobot-v1 using 5 random seeds and 1000 episodes per setting. Among the
tested primary hyperparameters, $T$ showed the clearest task dependence.
CartPole-v1 preferred shorter update windows, whereas Acrobot-v1 preferred
larger windows, although the latter also produced substantially larger final
network sizes. By contrast, $\theta_{\mathrm{var}}$ primarily controlled the
compactness of the final architecture: increasing
$\theta_{\mathrm{var}}$ markedly reduced final network size on both tasks
while only modestly changing late-training reward. The sensitivity trends for
$\lambda$ and $p_{\mathrm{rand}}$ were weaker and did not exhibit an equally
clear cross-task pattern.

Based on these results, the final shared defaults in
Table~\ref{tab:spm_hyperparams} were chosen as a practical cross-task
compromise rather than as a single-task optimum. In particular, $T=100$ was
retained as an intermediate setting, and $\theta_{\mathrm{var}}=0.1$ was
retained as a balanced compactness setting.

\begin{figure*}[t]
    \centering
    \begin{subfigure}[b]{0.48\textwidth}
        \centering
        \includegraphics[width=\textwidth]{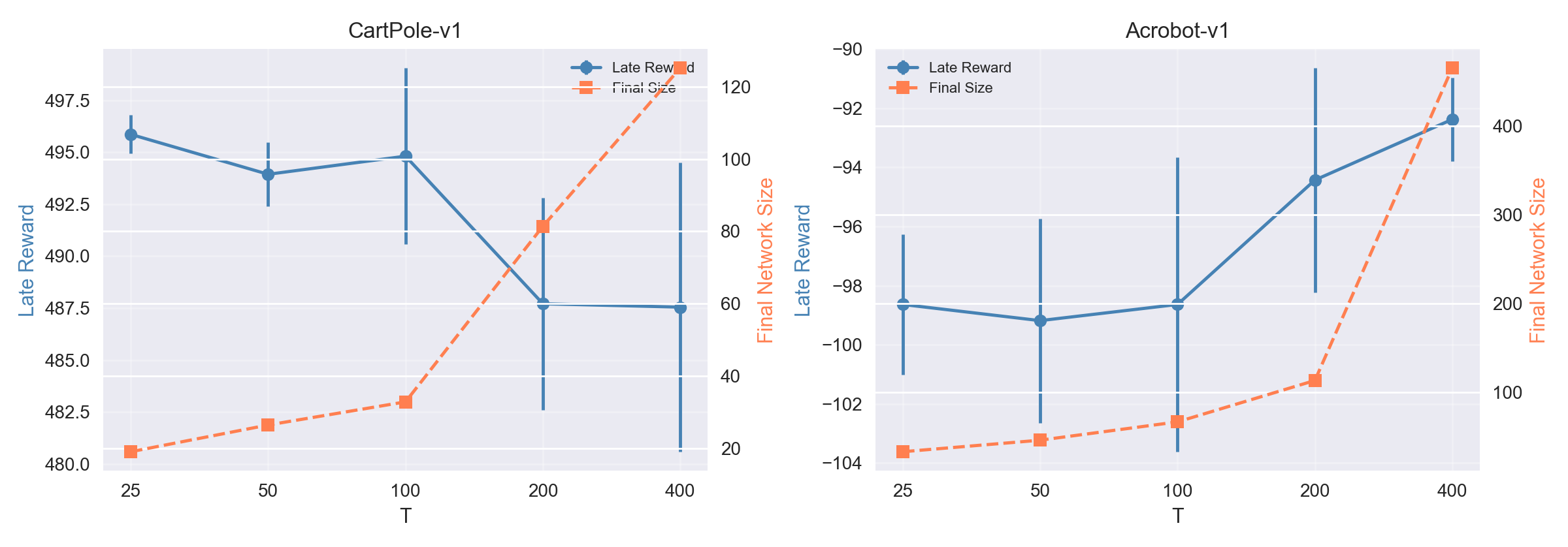}
        \caption{Sensitivity to $T$}
    \end{subfigure}
    \hfill
    \begin{subfigure}[b]{0.48\textwidth}
        \centering
        \includegraphics[width=\textwidth]{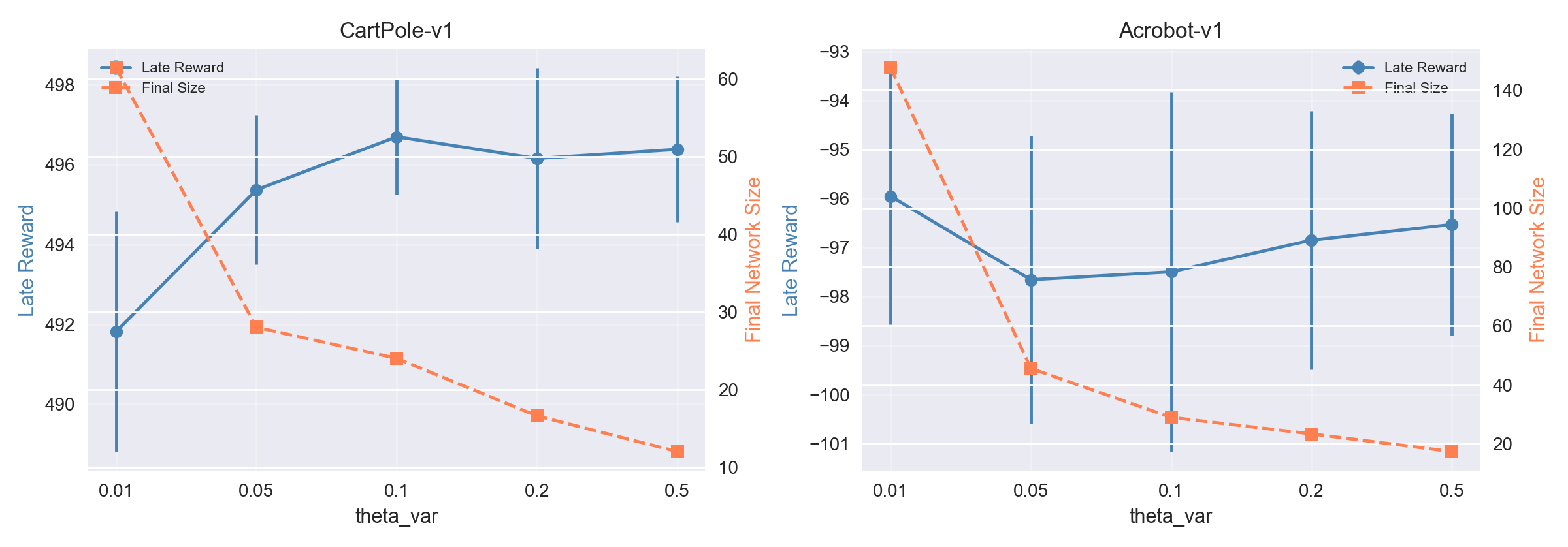}
        \caption{Sensitivity to $\theta_{\mathrm{var}}$}
    \end{subfigure}

    \vspace{0.5em}

    \begin{subfigure}[b]{0.48\textwidth}
        \centering
        \includegraphics[width=\textwidth]{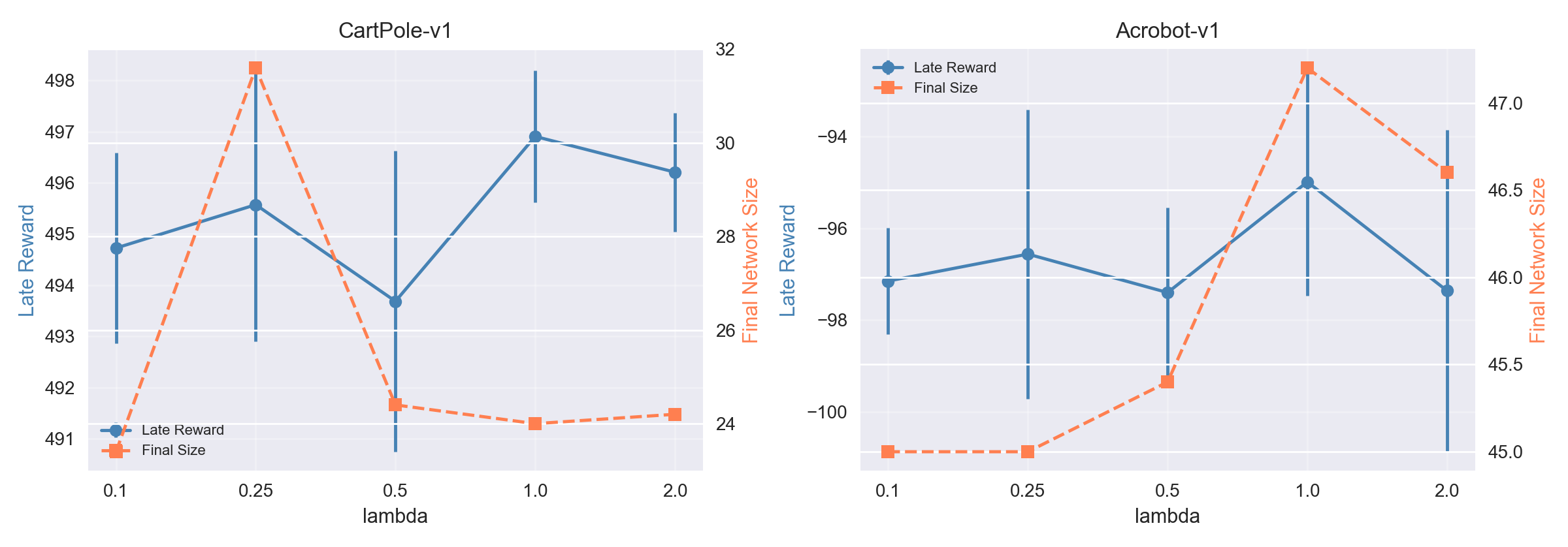}
        \caption{Sensitivity to $\lambda$}
    \end{subfigure}
    \hfill
    \begin{subfigure}[b]{0.48\textwidth}
        \centering
        \includegraphics[width=\textwidth]{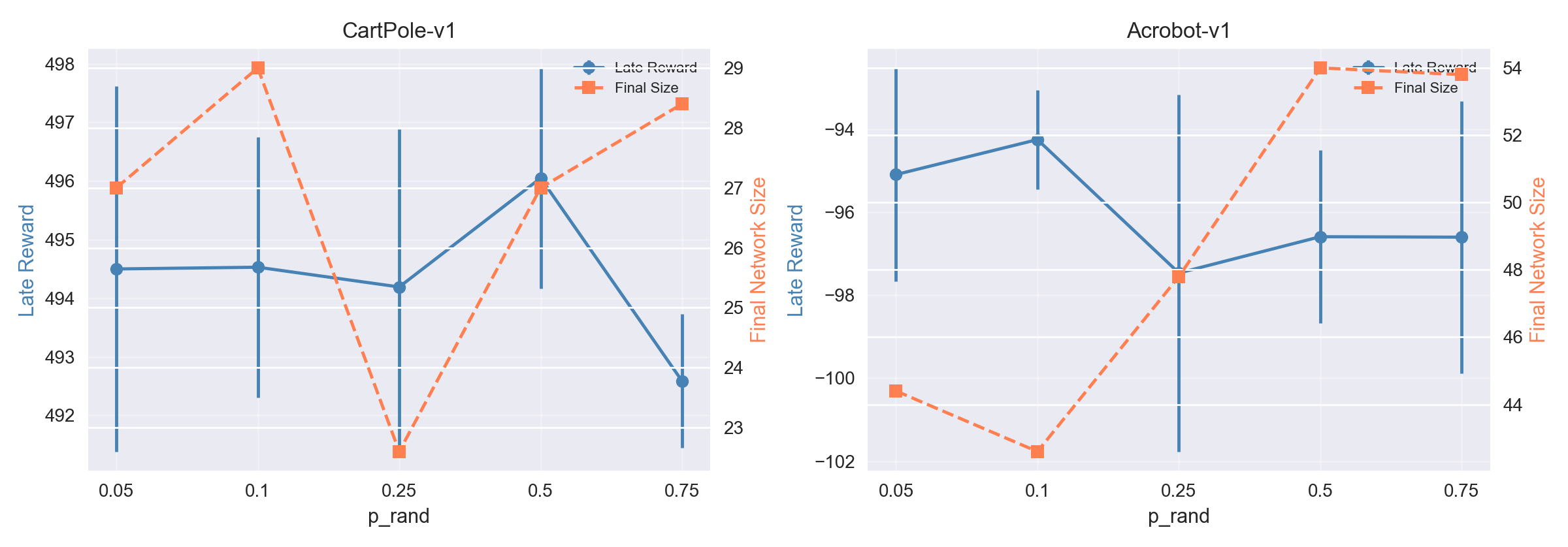}
        \caption{Sensitivity to $p_{\mathrm{rand}}$}
    \end{subfigure}
    \caption{One-factor-at-a-time sensitivity curves for the four primary SPM
    hyperparameters on CartPole-v1 and Acrobot-v1. Error bars indicate
    across-seed variability. Across the tested parameters, $T$ exhibits the
    clearest task dependence, while $\theta_{\mathrm{var}}$ most directly
    controls final network compactness.}
    \label{fig:hparam_sensitivity_appendix}
\end{figure*}

Fig.~\ref{fig:hparam_sensitivity_appendix} reports the one-factor-at-a-time
sensitivity curves for the four primary SPM hyperparameters. Consistent with
the summary above, $T$ shows the clearest task dependence, whereas
$\theta_{\mathrm{var}}$ mainly controls the compactness of the final network.
The effects of $\lambda$ and $p_{\mathrm{rand}}$ are comparatively weaker and
do not exhibit an equally clear cross-task pattern.

\subsection{Ablation on structural growth}
\label{app:ablation_growth_appendix}

\begin{table*}[t]
\centering
\caption{Ablation on structural growth in SMGrNN (mean $\pm$ std over 10 runs).
Late reward is the average return over the final quarter of episodes;
Conv.\ episodes is the number of episodes to reach the task-specific reward threshold;
Final size is the final number of hidden nodes.}
\label{tab:ablation_growth_stats}
\begin{tabular}{lcccc}
\hline
\textbf{Env.} & \textbf{Growth} & \textbf{Late reward} & \textbf{Conv.\ episodes} & \textbf{Final size} \\
\hline
CartPole    & Enabled  & $497.7 \pm 1.8$    & $147 \pm 15$ & 30.3 \\
            & Disabled & $489.8 \pm 12.1$   & $152 \pm 17$ & 6.0  \\
\hline
Acrobot     & Enabled  & $-94.0 \pm 1.9$    & $80 \pm 12$  & 47.5 \\
            & Disabled & $-95.9 \pm 5.5$    & $80 \pm 12$  & 9.0  \\
\hline
LunarLander & Enabled  & $207.6 \pm 33.3$   & $231 \pm 16$ & 48.8 \\
            & Disabled & $145.7 \pm 130.5$  & $234 \pm 38$ & 12.0 \\
\hline
\end{tabular}
\end{table*}

Table~\ref{tab:ablation_growth_stats} complements the main ablation
figure by reporting detailed statistics for the growth-enabled and
static variants. Across environments, enabling structural growth
consistently reduces variance and, on the more difficult tasks, yields
higher late-training rewards and larger final hidden layers than the
fixed-topology baseline.

\subsection{Additional growth-only weight distributions}
\label{app:growthonly_appendix_section}

\begin{figure*}[t]
    \centering
    \begin{subfigure}[b]{0.32\linewidth}
        \centering
        \includegraphics[width=\linewidth]{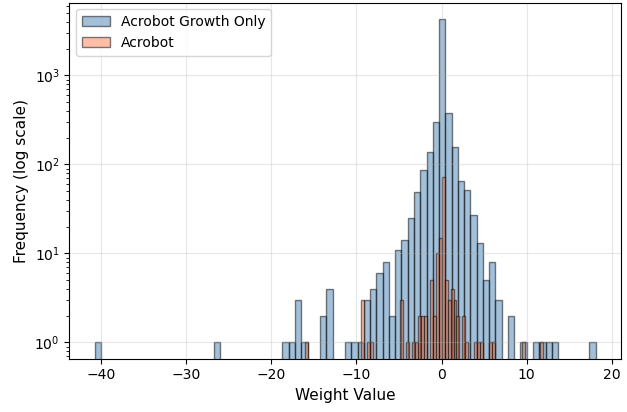}
        \caption{Acrobot log-scale hist.}
    \end{subfigure}
    \hfill
    \begin{subfigure}[b]{0.32\linewidth}
        \centering
        \includegraphics[width=\linewidth]{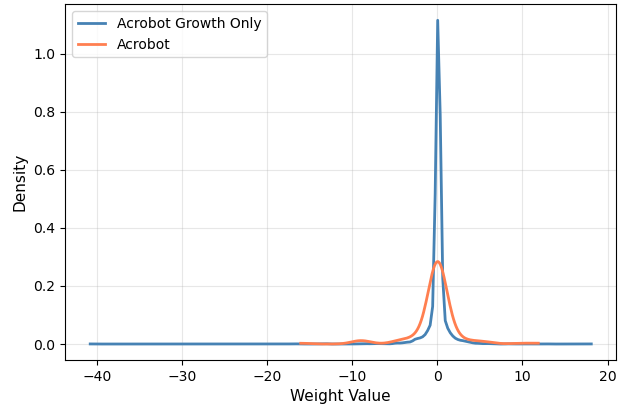}
        \caption{Acrobot PDF}
    \end{subfigure}
    \hfill
    \begin{subfigure}[b]{0.32\linewidth}
        \centering
        \includegraphics[width=\linewidth]{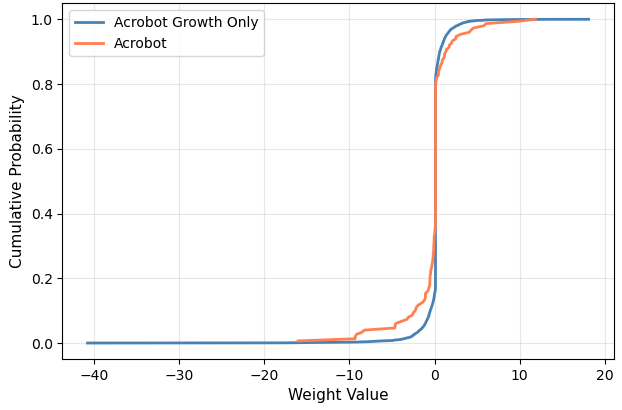}
        \caption{Acrobot CDF}
    \end{subfigure}

    \vspace{0.6em}

    \begin{subfigure}[b]{0.32\linewidth}
        \centering
        \includegraphics[width=\linewidth]{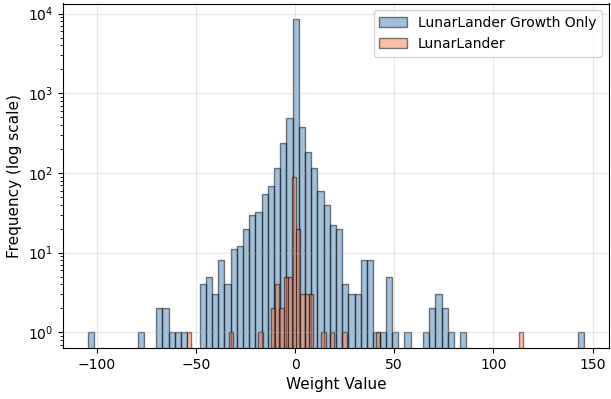}
        \caption{LunarLander log-scale hist.}
    \end{subfigure}
    \hfill
    \begin{subfigure}[b]{0.32\linewidth}
        \centering
        \includegraphics[width=\linewidth]{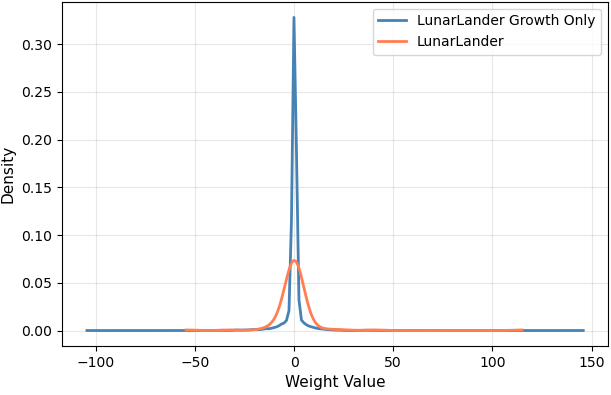}
        \caption{LunarLander PDF}
    \end{subfigure}
    \hfill
    \begin{subfigure}[b]{0.32\linewidth}
        \centering
        \includegraphics[width=\linewidth]{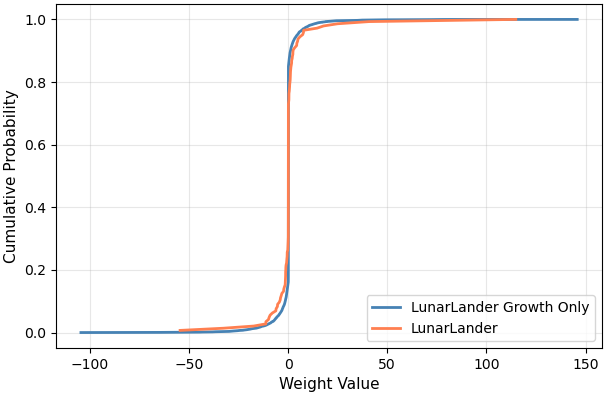}
        \caption{LunarLander CDF}
    \end{subfigure}

    \caption{Additional weight distributions for the growth-only and
    SMGrNN configurations on Acrobot-v1 and LunarLander-v3.}
    \label{fig:growthonly_appendix}
\end{figure*}

Fig.~\ref{fig:growthonly_appendix} extends the growth-only analysis to
Acrobot-v1 and LunarLander-v3. Log-scale histograms, PDFs, and CDFs show
that the growth-only variant produces very broad weight spectra with many
low-magnitude connections, whereas SMGrNN concentrates weight mass on a
more compact set of edges, consistent with the effect of pruning.

\subsection{Initial-topology sensitivity}
\label{app:topology_appendix}

\begin{table*}[t]
\begin{minipage}[t]{0.62\textwidth}
\centering
\caption{Initial-topology sensitivity of SMGrNN (mean $\pm$ std over 10 runs).
Late reward is the average return over the final quarter of episodes;
Conv.\ episodes is the number of episodes to reach the task-specific
reward threshold; Net growth is the average increase in hidden-node
count relative to the initial topology.}
\label{tab:topology_summary}
\begin{tabular}{lcccc}
\hline
\textbf{Env.} & \textbf{Init nodes} & \textbf{Late reward} & \textbf{Conv.\ episodes} & \textbf{Net growth} \\
\hline
CartPole    & 10 & $495.2 \pm 2.6$   & $142 \pm 14$ & 21.9 \\
            & 5  & $496.1 \pm 1.7$   & $153 \pm 18$ & 22.1 \\
            & 0  & $497.7 \pm 1.8$   & $147 \pm 15$ & 24.2 \\
\hline
Acrobot     & 10 & $-95.0 \pm 2.2$   & $84 \pm 12$  & 31.1 \\
            & 5  & $-95.0 \pm 3.6$   & $82 \pm 12$  & 33.7 \\
            & 0  & $-94.0 \pm 1.9$   & $80 \pm 12$  & 35.4 \\
\hline
LunarLander & 10 & $232.2 \pm 31.2$  & $231 \pm 27$ & 29.8 \\
            & 5  & $222.4 \pm 34.1$  & $240 \pm 21$ & 32.5 \\
            & 0  & $207.6 \pm 33.3$  & $231 \pm 16$ & 34.0 \\
\hline
\end{tabular}
\end{minipage}
\hfill
\begin{minipage}[t]{0.35\textwidth}
\centering
\caption{Final network size and net growth vs.\ task difficulty
(mean $\pm$ std over 10 runs).
Final size is the number of hidden nodes at the end of training;
Net growth is the average increase in hidden-node count relative to the
initial topology.}
\label{tab:difficulty_summary}
\begin{tabular}{lcc}
\hline
\textbf{Task} & \textbf{Final size} & \textbf{Net growth} \\
\hline
XOR            & $11.6 \pm 2.0$  & 7.9  \\
CartPole-v1    & $30.3 \pm 5.6$  & 24.2 \\
Acrobot-v1     & $47.5 \pm 9.0$  & 35.4 \\
LunarLander-v3 & $48.8 \pm 14.0$ & 34.0 \\
\hline
\end{tabular}
\end{minipage}
\end{table*}

Table~\ref{tab:topology_summary} reports detailed statistics for the
initial-topology sensitivity experiment. Across initial hidden sizes in
$\{0,5,10\}$, late rewards and convergence episodes remain very similar,
while net growth compensates for the initial capacity: sparser initial
graphs exhibit slightly larger net increases in hidden-node count. These
results support the conclusion that SMGrNN is relatively robust to the
choice of initial hidden-layer size.

\subsection{Task-difficulty scaling}
\label{app:difficulty_appendix}

Table~\ref{tab:difficulty_summary} summarizes how emergent network size
depends on task difficulty. The trivial XOR task stabilizes at the smallest
hidden layer, CartPole-v1 converges to a modest size, and Acrobot-v1 and
LunarLander-v3 yield the largest final networks and net growth. This
ordering mirrors the qualitative difficulty of the tasks and reinforces the
observation that, under fixed SPM hyperparameters, SMGrNN allocates
capacity in proportion to task demands.

\end{document}